\documentclass{article}

\usepackage{microtype}
\usepackage{graphicx}
\usepackage{subcaption}
\usepackage{booktabs} 

\usepackage[accepted]{stylefile}

\usepackage{amsmath}
\usepackage{amssymb}
\usepackage{mathtools}
\usepackage{amsthm}

\theoremstyle{plain}

\theoremstyle{definition}

\theoremstyle{remark}

\usepackage[textsize=tiny]{todonotes}

\usepackage{balance}

\usepackage[utf8]{inputenc} 
\usepackage[T1]{fontenc}    
\usepackage[colorlinks=true, allcolors=blue]{hyperref}
\usepackage{url}            
\usepackage{booktabs}       
\usepackage{amsfonts}       
\usepackage{nicefrac}       
\usepackage{microtype}      
\usepackage{xcolor}         

\usepackage[capitalize,noabbrev]{cleveref}

\usepackage{amsmath, amssymb, amsfonts, dsfont, mathabx,mathtools}

\usepackage{float}
\usepackage{wrapfig}
\usepackage{thmtools,thm-restate}

 \usepackage{placeins}

\usepackage{algorithmic}
\usepackage{algorithm}
\usepackage{graphicx}
\usepackage{float}
\usepackage{csvsimple}
\usepackage{enumitem} 

\renewcommand\cite{\citep} 

\DeclareMathOperator*{\argmax}{arg\,max}

\newcommand{\beginsupplement}{ 
        \setcounter{section}{0}
        \setcounter{table}{0}
        \renewcommand{\thetable}{S\arabic{table}} %
        \setcounter{figure}{0}
        \renewcommand{\thefigure}{S\arabic{figure}} %
     }

\newcommand{\methodabbrev}{CROWDLAB}

\newcommand{\appendixtitle}{Appendix -- CROWDLAB: Supervised learning to infer consensus labels \\ \hspace*{1.8em} and quality scores for data with multiple annotators}

\newcommand{\gitlink}{\url{https://github.com/cleanlab/multiannotator-benchmarks}}

\icmltitlerunning{CROWDLAB: Supervised learning to infer consensus labels and quality scores for data with multiple annotators}

\begin{document}

\twocolumn[
\icmltitle{CROWDLAB: Supervised learning to infer consensus labels \\ and quality scores for data with multiple annotators}



\icmlsetsymbol{equal}{*}

\begin{icmlauthorlist}
\icmlauthor{Hui Wen Goh}{comp}
\icmlauthor{Ulyana Tkachenko}{comp}
\icmlauthor{Jonas Mueller}{comp}
\end{icmlauthorlist}

\icmlaffiliation{comp}{Cleanlab}

\icmlcorrespondingauthor{HWG}{huiwen@cleanlab.ai}
\icmlcorrespondingauthor{UT}{ulyana@cleanlab.ai}
\icmlcorrespondingauthor{JM}{jonas@cleanlab.ai}

\icmlkeywords{Machine Learning, ICML}

\vskip 0.3in
]



\printAffiliationsAndNotice{} 

\begin{abstract}
Real-world data for classification is often labeled by multiple annotators. For analyzing such data, we introduce \methodabbrev{}, a straightforward approach to utilize \emph{any} trained classifier to estimate: 
(1) A consensus label for each example that aggregates the available annotations;  
(2) A confidence score for how likely each consensus label is correct; 
(3) A rating for each annotator quantifying the overall correctness of their labels. 
Existing algorithms to estimate related quantities in crowdsourcing often rely on sophisticated generative models with iterative inference. \methodabbrev{} instead uses a straightforward  weighted ensemble. Existing algorithms often rely solely on annotator statistics, ignoring the features of the examples from which the annotations derive. \methodabbrev{} utilizes \emph{any} classifier model trained on these features, and can thus better generalize between examples with similar features. On real-world multi-annotator image data, our proposed method provides superior estimates for (1)-(3) than  existing algorithms like Dawid-Skene/GLAD.
\end{abstract}

\section{Introduction}

Training data for multiclass classification are often labeled by multiple annotators, with some redundancy between annotators to ensure high-quality labels. Such settings have been studied in crowdsourcing research \cite{munro2021human, paun2018comparing}. There it is often assumed that \emph{many} annotators have labeled each example \cite{carpenter2008multilevel, khetan2018learning}, but this can be prohibitively expensive. This paper considers general settings where each example in the dataset is merely labeled by at least \emph{one} annotator, and each annotator labels many examples (but still only a subset of the dataset). 
Each annotation corresponds to the selection of one class $y \in \{1,\dots, K\}$ which the annotator believes to be most appropriate for this example. 

Certain classification models can be trained in a special manner to account for the multiple labels per example \cite{nguyen2014learning, cifar10h}, but this is rarely done in practical applications. A common approach is to aggregate the labels for each example into a single \emph{consensus label}, e.g.\ via majority-vote or statistical crowdsourcing algorithms \cite{dawidskene}. Any classifier can then be trained on these consensus labels via off-the-shelf code. 

Here we propose a method\footnote{
Code:  \url{https://github.com/cleanlab/cleanlab} \\ Reproduce our  results: \gitlink{}}
that leverages any already-trained classifier to: (1) establish accurate consensus labels, (2) estimate their quality, and (3) estimate the quality of each annotator \cite{qualitycontrol}. The latter two aims help us determine which data is least trustworthy and should perhaps be verified via  additional annotation \cite{bernhardt2022active}. \textbf{\methodabbrev{}} (Classifier Refinement Of croWDsourced LABels) is based on a straightforward weighted ensemble of the classifier predictions and individual annotations. Weights are assigned according to the (estimated) trusthworthiness of each annotator relative to the trained classifier.  \methodabbrev{} is easy to implement/understand, computationally efficient (non-iterative), and extremely flexible. It works with \emph{any} classifier and training procedure, as well as \emph{any} classification dataset (including those containing examples only labeled by one annotator).  

\begin{figure*}
\begin{subfigure}[b]{.5\textwidth} 
  \includegraphics[width=0.745\textwidth]{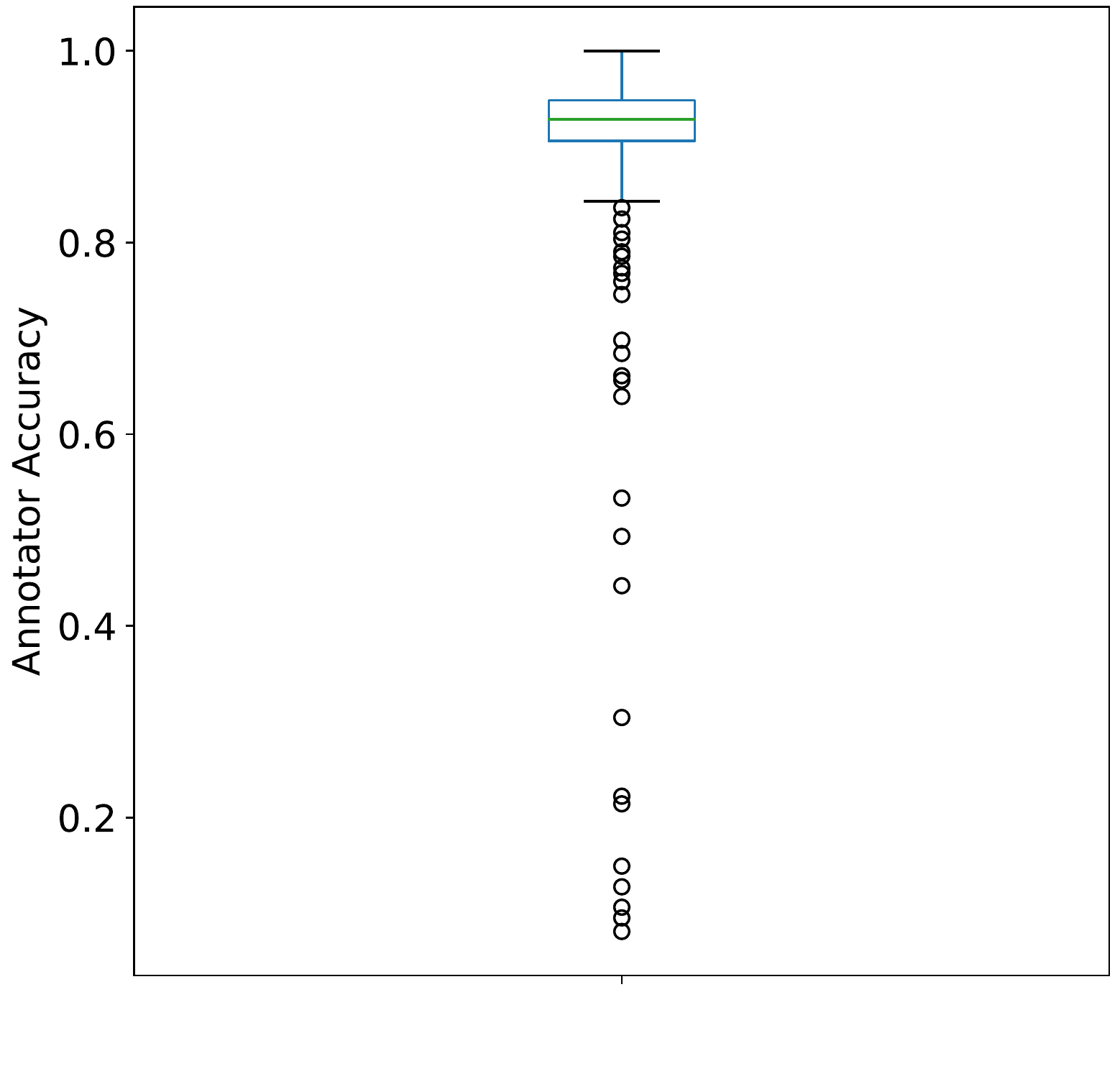} 
  \label{fig:annotator}
\end{subfigure}
 \hspace*{0.06\textwidth}
\begin{subfigure}[b]{.5\textwidth} 
  \includegraphics[width=0.77\textwidth]{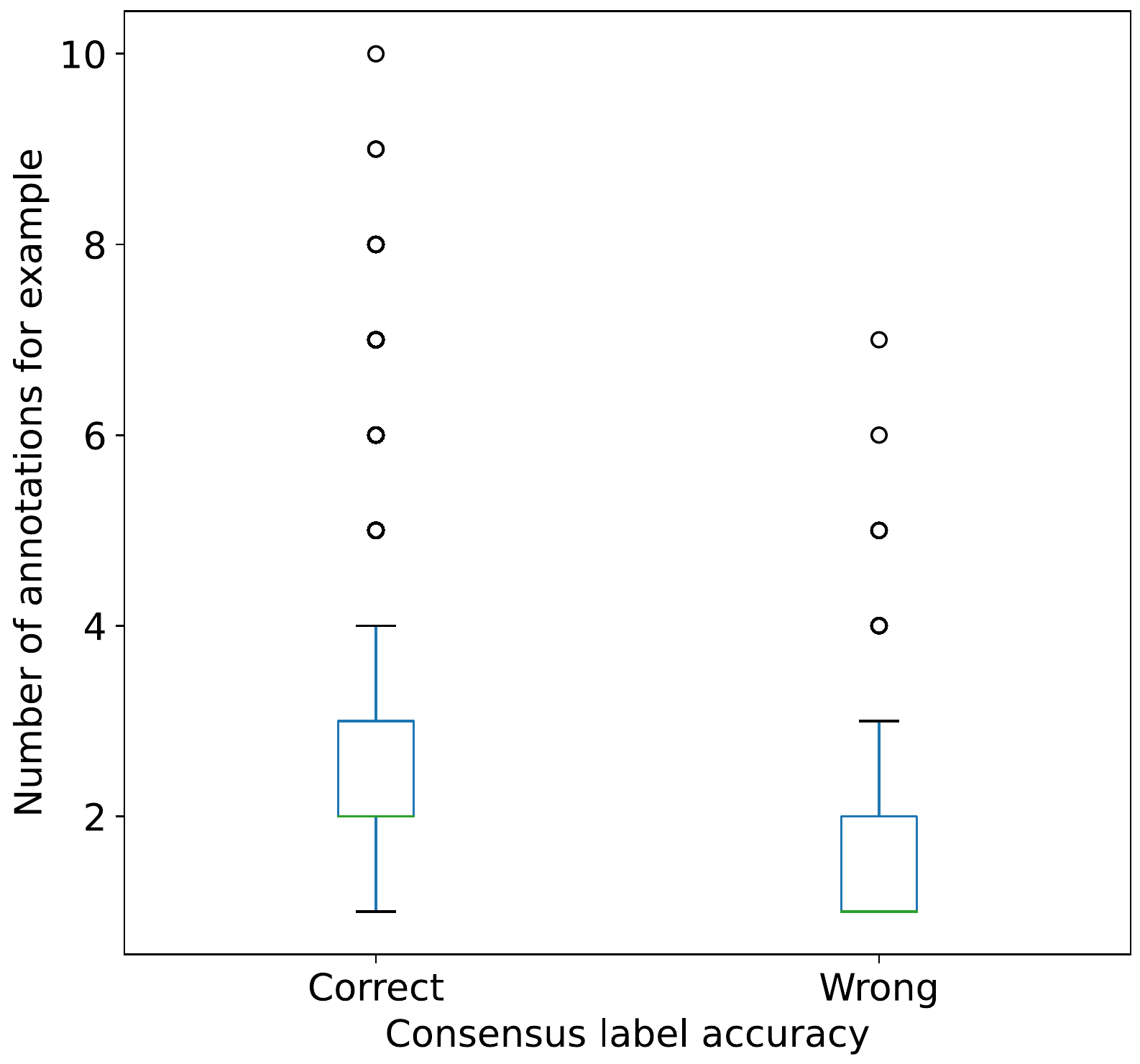} 
  \label{fig:consensus}
\end{subfigure}
\vspace*{-6mm}
\caption{Statistics of our \emph{Hardest} dataset, which has  images annotated by many actual humans. \textbf{(a)} Distribution over annotators showing the overall accuracy of each annotator's chosen labels. 
\textbf{(b)} Distribution over examples showing the number of annotations per example, grouped by whether the majority-vote consensus label is correct or not.  Accuracy is measured  against underlying ground-truth labels. 
}
\label{fig:motivation}
\end{figure*}

\paragraph{Motivations.} Illustrating how many real-world multi-annotator datasets look, Figure \ref{fig:motivation} shows a disparity in annotator quality as well as many examples whose consensus label will be incorrect if we rely on majority vote (nonetheless often done in practice due to its straightforward appeal). Unsurprisingly, consensus labels are more likely to be incorrect for those examples with fewer annotations. An effective method to estimate consensus label quality should properly account for the number of annotations an example has received, as well as the quality of the annotators who selected these labels. Many of the examples whose consensus label is wrong merely have a single annotation, which provides little information. Using a trained classifier can help us better generalize to such examples to estimate their labels' quality (especially if the data contain other examples with similar feature values). When incorporating a classifier, we also wish to account for the accuracy and confidence of its estimates. CROWDLAB offers a straightforward way to appropriately account for all of these factors.

\section{Methods}
\label{sec:methods}

\newcommand{\probclassifier}[1]{\widehat{p}_{\mathcal{M}}(Y_{#1} \mid X_{#1} )}
\newcommand{\probclassifierk}[1]{\widehat{p}_{\mathcal{M}}(Y_{#1} = k \mid X_{#1} )}
\newcommand{\probclassifiersim}[1]{\widehat{p}_{\mathcal{M}}(Y_{#1} \mid X_{#1} )}
\newcommand{\probclassifiershort}{\widehat{p}_{\mathcal{M},i,k}}
\newcommand{\probsomething}[2]{\widehat{p}_{#2}(Y_{#1} \mid X_{#1} )} 
\newcommand{\probsomethingk}[2]{\widehat{p}_{#2}(Y_{#1} = k \mid X_{#1} )}
\newcommand{\probempk}[1]{\widehat{p}_\text{emp}(Y_{#1} = k \mid \{ Y_{ij} \}_{j \in \mathcal{J}_i} )}
\newcommand{\posterior}[2]{\widehat{p}_{\text{#2}}(Y_{#1} \mid X_{#1}, \{ Y_{ij} \} )}
\newcommand{\posteriornox}[2]{\widehat{p}_{\text{#2}}(Y_i \mid \{ Y_{ij} \} )}
\newcommand{\posteriornoxnotxt}[2]{\widehat{p}_{#2}(Y_i \mid \{ Y_{ij} \} )}
\newcommand{\posteriornoxnotxtk}[2]{\widehat{p}_{#2}(Y_i = k \mid \{ Y_{ij} \} )}
\newcommand{\predictedlabel}{Y_{i,\mathcal{M}}}

Consider a dataset sampled from (feature, class label) pairs $(X,Y)$ that is comprised of: $n$ examples, $K$ classes, and $m$ annotators in total. We first establish some notation to formally describe our setting:  
$| \mathcal{J} |$ denotes the cardinality of set $\mathcal{J}$, $\mathds{1} (\cdot)$ is an indicator function emitting 1 if its condition is True and 0 otherwise, and 
$[n] = \{ 1,2,..., n \}$ indexes examples in the dataset. $X_i$ denotes the features of the $i$th example, which belongs to one class $Y_i \in [K]$. This true class is unknown to us.  
For $j \in [m]$: $\mathcal{A}_j$ denotes the $j$th annotator, and $Y_{ij} \in [K]$ is the class this annotator chose for $X_i$. $Y_{ij} = \emptyset$ if $\mathcal{A}_j$ did not label this particular example. Each example receives at most $m$ annotations, with many examples receiving fewer. 
$\widehat{Y}_i$ denotes the consensus label for example $i$, representing our best estimate of its true class $Y_i$. 
$\mathcal{I}_j := \{ i \in [n] : Y_{ij} \neq \emptyset \}$ denotes the subset of examples labeled by $\mathcal{A}_j$. We assume each annotator has labeled multiple examples,  i.e.\ $|\mathcal{I}_j| > 1$. 
$\mathcal{J}_i := \{ j \in [m] : Y_{ij} \neq \emptyset \} $ denotes the subset of annotators that labeled $X_i$. Some examples may only be labeled by a single annotator. 



We assume some classifier model $\mathcal{M}$ has been trained to predict the given labels based on feature values. \methodabbrev{} can be used with any type of classifier $\mathcal{M}$ (and training procedure), as long as it outputs   predicted class probabilities $\probclassifier{} \in \mathbb{R}^K$ estimating the likelihood that example $X$ belongs to each class $k$. 
To avoid overfit  predictions, we fit $\mathcal{M}$ via cross-validation. This provides  \emph{held-out} predictions $\probclassifier{i}$ for each example in the dataset (from a copy of $\mathcal{M}$ which never saw $X_i$ during training). 
In our experiments, we simply train $\mathcal{M}$ on consensus labels derived via majority vote. But one could train the classifier on any other set of consensus labels or even on the individual labels from each annotator (simply duplicating multiply-annotated examples in the training set). 
All methods considered here that use $\mathcal{M}$ will benefit from improvements in the classifier's predictive accuracy. However  \methodabbrev{} is the only method that  explicitly accounts for shortcomings of the classifier's predictions (inevitable due to estimation error). 

\subsection{Scoring Consensus Quality}
\label{sec:cqs}

We first outline methods to estimate our confidence that a given consensus label for each example is correct. These quality estimates $q_i \in [0,1]$ may be applied to any given label no matter which method was used to establish consensus. 
Once we can estimate the quality of any one label for each example, we estimate the best consensus label under each method as the class with the highest consensus quality score. This class can be identified efficiently for CROWDLAB. 
CROWDLAB combines the complementary strengths of two basic estimators that we discuss first.

\textbf{Agreement} \cite{munro2021human}. \ 
The fraction of annotators who agree with consensus label (does not use a classifier).
\begin{equation}
    q_i = \frac{1}{| \mathcal{J}_i |} \sum_{j \in \mathcal{J}_i} \mathds{1} ( Y_{ij} = \widehat{Y}_i )
\vspace*{-1mm}
\end{equation}
Final consensus labels can be established via majority vote.

\textbf{Label Quality Score} \cite{labelerrordetection}. \ 
Instead of relying on the annotators, one can rely on the classifier model via methods used to evaluate labels in standard (singly-labeled) classification datasets. This approach ignores information from individual annotators when computing consensus quality: 
$\displaystyle q_i = L(\widehat{Y}_i, \probclassifier{i} )$.

Here $L(Y, \hat{p}) \in [0,1]$ is a \emph{label quality score}  which quantifies our confidence that a particular label $Y \in [K]$ is correct for example $X$, given model-prediction $\hat{p} \in \mathbb{R}^K$ estimating the likelihood that $X$ belongs to each class. Our work uses \emph{self-confidence} as the label quality score: $L(Y, p) = \hat{p}(Y \mid X)$. This simply  represents the model-estimated probability that the example belongs to its labeled class. \citet{labelerrordetection, northcutt2021confidentlearning} found this to be effective for scoring label errors in singly-labeled data based on classifier predictions.

\textbf{CROWDLAB (Classifier Refinement Of croWDsourced LABels)}. 
The aforementioned approaches fail to consider both annotators and  classifier. Treating these as different predictors of an example's true label, we take inspiration from prediction competitions where weighted ensembling of predictors produces  accurate and calibrated predictions. 
CROWDLAB also employs the same label quality score for each consensus label, but applies it to a different class probability vector which modifies the prediction output by our classifier to account for the individual annotations for an example: 
$\displaystyle q_i = L(\widehat{Y}_i, \posterior{i}{\text{CR}})$. \looseness=-1

We estimate these class probabilities by means of a weighted ensemble aggregation \cite{fakoor2021flexible}:
\begin{align*}
    & \posterior{i}{\text{CR}} = \nonumber \\
    & \quad \frac{w_{\mathcal{M}} \cdot \probclassifiersim{i} + \sum_{j \in \mathcal{J}_i} w_j \cdot  \posteriornoxnotxt{i}{\mathcal{A}_j}}{w_{\mathcal{M}} + \sum_{j \in \mathcal{J}_i} w_j}
    \label{eq:our_cqs}
\end{align*}
Here $\widehat{p}_{\mathcal{M}} \in \mathbb{R}^K$ is the probability of each class predicted by our classifier, $\widehat{p}_{\mathcal{A}_j} \in \mathbb{R}^K$ is a similar likelihood vector treating each annotator's label as a probabilistic ``prediction'', and $w_j, w_{\mathcal{M}} \in \mathbb{R}$ are weights to account for the (estimated) relative trustworthiness of each annotator and our classifier.
Our estimation procedure for these weights ensures  $w_{\mathcal{M}}$ is smaller  if our classifier was poorly trained and $w_j$ is smaller for the annotators who give less accurate labels overall.

To present the remaining details, we first define a likelihood parameter $P$ as the average annotator agreement, across examples with more than one annotation. $P$ estimates the probability that an arbitrary annotator's label will match the majority-vote consensus label for an arbitrary example.
\begin{gather}
    P = \frac{1}{|\mathcal{I}_+|} \sum_{i \in \mathcal{I}_+} \frac{1}{| \mathcal{J}_i |} \sum_{j \in \mathcal{J}_i} \mathds{1} ( Y_{ij} = \widehat{Y}_i ) \ \nonumber \\
    \text{ where } \mathcal{I}_+ := \{i \in [n] : |\mathcal{J}_i| > 1 \}
    \label{eq:morethanone}
\end{gather}
We then simply define our per annotator predicted class probability vector used in (\ref{eq:our_cqs}) to be:
\begin{equation}
    \posteriornoxnotxtk{i}{\mathcal{A}_j} = \begin{cases}
    P & \mbox{when } Y_{ij} = k\\
    \frac{1 - P}{K - 1} & \mbox{when } Y_{ij} \neq k
\end{cases}
\label{eq:annotator_likelihood}
\end{equation}
This likelihood is shared across annotators and only involves a single parameter $P$, easily estimated from limited data. $P$ is a simple estimate of the accuracy of labels from a typical annotator. 
Note that including singly-annotated examples in (\ref{eq:morethanone}) would bias $P$. 
This likelihood facilitates comparing classifier outputs against outputs from the typical annotator. 

Now we detail how to estimate the trustworthiness weights $w_j, w_{\mathcal{M}}$.
Let $s_j$ represent annotator $j$'s agreement with other annotators who labeled the same examples:
\begin{equation}
    s_j = \frac{\sum_{i \in \mathcal{I}_j} \sum_{\ell \in \mathcal{J}_i, \ell \neq j} \mathds{1} (Y_{ij} = Y_{i\ell})}{\sum_{i \in \mathcal{I}_j} ( |\mathcal{J}_i| -1 )}
    \label{eq:sj}
\end{equation}
Let $A_{\mathcal{M}}$ be the (empirical) accuracy of our classifier with respect to the majority-vote consensus labels over the examples with more than one annotation:
\begin{equation}
    A_{\mathcal{M}} = \frac{1}{|\mathcal{I}_+|} \sum_{i \in \mathcal{I}_+} \mathds{1} (\predictedlabel{} = \widehat{Y_i})  \label{eq:am} 
    \end{equation}
Here $\predictedlabel{} :=  \argmax_k \probclassifierk{i} \in [K]$ is the class predicted by our model for $X_i$.
$A_{\mathcal{M}}$ and $s_j$ from (\ref{eq:sj}) are analogous accuracy estimates for our classifier and individual annotators. Both are computed with only the multiply-annotated examples, since majority-vote consensus labels are more reliable for this subset.

Before defining the trustworthiness weights, we normalize these accuracy estimates with respect to a baseline that puts them on a  meaningful scale. This baseline is based on the estimated accuracy $A_{\text{MLC}}$ of always predicting the overall \emph{most labeled class} across all examples'  annotations $Y_{\text{MLC}} := \argmax_k \sum_{ij} \mathds{1} (Y_{ij} = k)$, i.e.\ the class selected the most by the annotators across all examples. This accuracy is also estimated on only the subset of examples that have more than one annotator, $\mathcal{I}_+$ defined in (\ref{eq:morethanone}). 
\begin{equation}
    A_{\text{MLC}} = \frac{1}{|\mathcal{I}_+|} \sum_{i \in \mathcal{I}_+} \mathds{1} (Y_{\text{MLC}} = \widehat{Y_i})
\end{equation}
Adopting this most-labeled-class-accuracy as a baseline, we compute normalized versions of our estimates for: each annotator's agreement with other annotators and the adjusted accuracy of the model.
\begin{equation}
    w_j = 1 - \frac{1 - s_j}{1 - A_{\text{MLC}}}
    \label{eq:annotatorweight}
\end{equation}
\begin{equation}
    w_{\mathcal{M}} = \left( 1 - \frac{1 - A_{\mathcal{M}}}{1 - A_{\text{MLC}}} \right) \cdot \sqrt{\frac{1}{n} \sum_i |\mathcal{J}_i|}
    \label{eq:modelweight}
\end{equation}
CROWDLAB uses $w_j$ and $w_{\mathcal{M}}$ to weight our annotators and classifier model in its weighted ensemble of predictors. Each trustworthiness weight can thus be understood as 1 minus the (estimated) relative error of the corresponding predictor. Such normalized-error based weighting is commonly employed to combine predictors  in \emph{model averaging}.


\subsection{Scoring Annotator Quality}
\label{sec:aqs}

Beyond estimating consensus labels and their quality, we consider ranking which annotators provide the best/worst labels. Here are methods to get an overall quality score $a_j \in [0,1]$ summarizing each annotator's accuracy/skill.  

\textbf{Agreement} \cite{munro2021human}. \ 
A simple score is the empirical accuracy of each annotator's labels with respect to majority-vote consensus labels. Examples with one annotation are not considered in this  calculation to reduce bias.
\begin{gather}
    a_j = \frac{1}{| \mathcal{I}_{j,+} |} \sum_{i \in \mathcal{I}_{j,+}} \mathds{1} ( Y_{ij} = \widehat{Y}_i ) \ \ \nonumber \\
    \text{ where  } \mathcal{I}_{j,+} := \mathcal{I}_j \cap  \mathcal{I}_+ = \{i \in \mathcal{I}_j : |\mathcal{J}_i| > 1 \}
    \label{eq:agreement_annotator_score}
\end{gather}

\textbf{Label Quality Score} \cite{labelerrordetection}. \ 
Agreement scores rate annotators solely based on labeling statistics. We can also rely on our classfier predictions $\widehat{p}_{\mathcal{M}}$ to rate the average quality of all labels provided by one annotator. 
\begin{equation}
    a_j = \frac{1}{|\mathcal{I}_j|} \sum_{i \in \mathcal{I}_j} L \big(Y_{ij}, \probclassifier{i} \big) 
    \label{eq:lqs_annotator}
\end{equation}

\textbf{CROWDLAB}. \ 
Our method takes into account both the label quality score of each annotator's labels (computed based on our classifier), as well as the agreement between each annotator's label and the CROWDLAB consensus label. 
As in (\ref{eq:lqs_annotator}), we estimate an  average label quality score of labels given by each annotator, but here using estimated class probabilities $\widehat{p}_{\text{CR}}$ from   CROWDLAB in Sec. \ref{sec:cqs}:
\begin{equation}
    Q_j = \frac{1}{|\mathcal{I}_j|} \sum_{i \in \mathcal{I}_j} L \big( Y_{ij}, \posterior{i}{CR} \big) 
    \label{eq:qj}
\end{equation}
Next, we compute each annotator's agreement with consensus among examples with over one annotation.
\begin{gather}
    A_j = \frac{1}{| \mathcal{I}_{j,+} |} \sum_{i \in \mathcal{I}_{j,+}} \mathds{1} ( Y_{ij} = \widehat{Y}_i ) 
    \label{eq:aj}
\end{gather}
Here $\mathcal{I}_{j,+}$ is defined in (\ref{eq:agreement_annotator_score}) and the consensus labels $\widehat{Y}_i$ are established via the CROWDLAB method from Sec. \ref{sec:cqs}.
Since CROWDLAB is an effective method to estimate consensus labels $\widehat{Y}_i$, one might wonder why $A_j$ alone from (\ref{eq:aj}) does not produce the best estimate of annotator quality. One reason is that $A_j$ fails to account for our \emph{confidence} in each consensus label and \emph{how} individual annotators deviate from consensus. If two annotators exhibit the same overall rate of agreement with the consensus labels, we should favor the annotator whose deviations from consensus are predicted to be likely classes by the classifier and tend to occur for examples with lower consensus quality score. 

Therefore we base CROWDLAB's annotator quality score on a weighted average between $A_j$ and $Q_j$. 
Using the model/annotator weights $w_\mathcal{M}, w_j$ computed by CROWDLAB in (\ref{eq:annotatorweight}) and (\ref{eq:modelweight})), we find a single aggregate weight to compare all annotators against the classifier.
\begin{gather}
   \widebar{w} = \frac{w_\mathcal{M}}{w_\mathcal{M} + w_0} \ \ \nonumber \\
   \text{ where } \ w_0 = \frac{1}{nm} \sum_{i=1}^n \sum_{j=1}^m w_j \cdot |\mathcal{J}_i|
\end{gather}
Here $\widebar{w}$ is shared across all annotators. It represents the (estimated) relative trustworthiness of our classifier against the average annotator. 
A quality score for each annotator is finally computed via a weighted average of: the label quality score and the annotator agreement with  consensus labels:
\begin{equation}
    a_j = \widebar{w} Q_j + (1-\widebar{w}) A_j
\end{equation}

\section{Related Work}
\label{sec:relwork}

Prior work for estimating (1)-(3) from multi-annotator datasets has fallen into two camps. 
The first camp relies on statistical generative models that only account for the observed annotator statistics \cite{carpenter2008multilevel}. Like CROWDLAB, the second camp of approaches also models feature-label relationships, but does so via linear models \cite{sideinfoglad}, autoencoders \cite{liu2021aggregating}, or classifiers fit to soft labels in an iterative manner \cite{raykar2010learning, khetan2018learning, rodrigues2018deep, platanios2020learning, liu2021aggregating}. These methods cannot utilize an arbitrary classifier (trained via any procedure), and they are more specialized and complex than CROWDLAB. 
Due to this complexity, approaches from the former camp  remain much more popular in practical applications \citep{crowdkit}. 

The following sections describe existing baseline methods to estimate consensus/annotator quality that our subsequent experiments compare CROWDLAB against. We focus our comparison on approaches which are either: commonly used in practice, or able to utilize \emph{any} classifier to produce better estimates (rather than approaches that are restricted to a specific type of model or non-standard training procedure). 

\subsection{Baseline Consensus Quality Scores}
\label{sec:cqsbase}

\textbf{Dawid-Skene} \cite{dawidskene}. \ 
This Bayesian method specifies a generative model of the dataset annotations. It employs iterative expectation-maximization (EM) to estimate each annotator's error rates in a class-specific manner. A key estimate in this approach is ${\posteriornox{i}{DS}}$, the posterior probability vector of the true class $Y_i$ for the $i$th example, given the dataset annotations $\{ Y_{ij} \}$. \looseness=-1 

Define $\pi_{k,\ell}^{(j)}$ as the probability that annotator $j$ labels an example as class $\ell$ when the true label of that example is $k$. This individual class confusion matrix for each annotator serves as the likelihood function of the Dawid-Skene generative model. 
The Dawid-Skene posterior distribution for a particular example is computed by taking product of each annotator's likelihood and some  prior distribution $\pi_{\text{prior}}$. \looseness=-1
\begin{equation}
    \posteriornox{i}{DS} \ \propto \ \ \pi_{\text{prior}} \cdot  \prod_{j \in \mathcal{J}_i} \pi_{k, Y_{ij}}^{(j)}
\end{equation}
Our work follows conventional practice taking the prior to be the (empirical) marginal distribution of given labels over the full dataset. A natural consensus quality score is the label quality score of the consensus label under the Dawid-Skene posterior class probabilities:    $\displaystyle q_i = L(\widehat{Y}_i,  \posteriornox{i}{DS})$.

\textbf{GLAD (Generative model of Labels, Abilities and Difficulties)} \cite{glad}. \ 
Specifying a more complex generative model of  dataset annotations than Dawid-Skene, this Bayesian approach also employs iterative EM steps. GLAD additionally infers $\alpha$, the expertise of each annotator and $\beta$, the difficulty of each example. GLAD's likelihood is based on the following  probability that an annotator chooses the same class as the consensus label:
\begin{equation}
    p (Y_{ij} = \widehat{Y}_i | \alpha_j, \beta_i) = \frac{1}{1 + e^{- \alpha_j \beta_i}}
\end{equation}
Like Dawid-Skene, GLAD uses the data likelihood to estimate the posterior probability of the true class $Y_i$ for the $i$th example:  $\posteriornox{i}{G}$. Here we use the same standard prior as for Dawid-Skene. 
Again a consensus quality score can naturally be obtained via the label quality score computed with respect to the GLAD posterior class probabilities: $ \displaystyle q_i = L(\widehat{Y}_i, \posteriornox{i}{G})$.

While other Bayesian annotation models exist  \cite{kara2015modeling, hovy2013learning, carpenter2008multilevel}, Dawid-Skene and GLAD are often used in practice \cite{crowdkit, qualitycontrol} and perform strongly in empirical benchmarks  \cite{sheshadri2013square,paun2018comparing,sinha2018fast}. \looseness=-1

\textbf{Dawid-Skene with Model} \cite{modelasannotator}. \ 
Although very popular, the Dawid-Skene and GLAD methods do not utilize a classifier at all. Thus they struggle with sparsely labeled examples. 
A straightforward adaptation of these methods to incorporate a classifier is to produce class predictions for each example (predict hard labels rather than probability vectors), and treat these predicted labels as if they were the outputs from an additional annotator \cite{modelasannotator}. Because methods like Dawid-Skene and GLAD automatically adjust for estimated annotator quality, they should theoretically 
account for the classifier's  strengths/weaknesses. 

We augment Dawid-Skene in this way by: adding the model's predicted labels as an additional annotator (for every example), and then computing consensus quality scores using the same Dawid-Skene method described above. 
The resulting posterior is now a function of the example's feature values as well (since classifier predictions depend on $X_i$).

\textbf{GLAD with Model} \cite{modelasannotator}. \ We follow the same approach to adapt GLAD to leverage the classifier: First add the model's predicted label for each example as labels from one additional annotator, and then compute the consensus quality score using the GLAD method described above.  \citet{sideinfoglad} proposed a different extension of GLAD to account for feature information, but their approach does not accommodate arbitrary classifiers (eg.\ not suitable for images). \looseness=-1

\textbf{Empirical Bayes.} \ 
While the previous two methods do not account for the classifier's confidence in its individual predictions, we consider an alternative adaptation of Dawid-Skene that does. 
This method treats the model's prediction as a per-example prior distribution and the annotators' labels as observations to compute $\posterior{i}{EB}$, the posterior probability of the true class $Y_i$ for the $i$th example, given the dataset annotations $\{ Y_{ij} \}$ and an example-specific prior based on the feature values $X_i$. 
The likelihood function for each annotator is defined by the class confusion matrix estimated via the Dawid-Skene algorithm. Using the classifier-derived prior distribution and likelihoods, we can compute an Empirical Bayes posterior in the same way outlined for Dawid-Skene:
\begin{equation}
    \posterior{i}{EB} \ \propto \ \ \probclassifiersim{i} \cdot  \prod_{j \in \mathcal{J}_i} \pi_{k, {Y}_{ij}}^{(j)}
\end{equation}
and compute a consensus quality score in the same manner:
$ \displaystyle    q_i = L(\widehat{Y}_i, \posterior{i}{EB} )$.

Some have considered iterative variants of this hybrid   generative/discriminative approach, in which the classifier is retrained to fit the resulting posterior and the above process is repeated with the new classifier \cite{raykar2010learning, khetan2018learning, rodrigues2018deep, platanios2020learning}. This however requires a classifier that can be iteratively trained over many rounds and also fit to soft labels, rather than a standard classification model.

\textbf{Active Label Cleaning} \cite{bernhardt2022active}. \ 
Also utilizing a trained classifier, this recently proposed method scores multi-annotator consensus quality by subtracting the  cross-entropy between classifier predicted probabilities and individual annotations by the entropy of the former. \looseness=-1
\begin{align}
    q_i = & - \sum_{k=1}^K \probempk{i} \cdot \log \probclassifiershort{}  \nonumber \\
    & - \left(  - \sum_{k=1}^K  \probclassifiershort{} \cdot \log \probclassifiershort{}  \right) 
\end{align}
Here we abbreviate $\probclassifiershort{} := \probclassifierk{i}$, and $\probempk{i}$ is the overall empirical distribution of class labels amongst the annotations for a particular example. 
Like CROWDLAB, this approach accounts for classifier confidence and all individual annotations. It lacks CROWDLAB's ability to adjust for how trustworthy the individual annotators and classifier are.

\subsection{Baseline Annotator Quality Scores}
\label{sec:aqsbase}

\textbf{Dawid-Skene} \cite{dawidskene}. \ 
We follow the conventional use of Dawid-Skene to rate a particular annotator via the probability that they agree with the true label. This is directly estimated for each possible true label as part of the per-annotator class confusion matrix used by the Dawid-Skene method  (see Sec.~\ref{sec:cqsbase}). Thus one can score each annotator using the trace of their confusion matrix.
\begin{equation}
    a_j = \frac{1}{K} \sum_{k=1}^K \pi_{k,k}^{(j)}
\end{equation}
\textbf{GLAD} \cite{glad}. \ 
Expertise of each annotator as estimated by GLAD method (see Sec.~\ref{sec:cqsbase}): \ 
 $\displaystyle   a_j = \alpha_j
$.

\textbf{Dawid-Skene with Model} \cite{modelasannotator}. \ 
Add the classfier's predicted labels as an additional annotator (who labeled every example).  Then score each real annotator's quality using the Dawid-Skene method above.

\textbf{GLAD with Model} \cite{modelasannotator}. \
Add the classifier's predicted labels as an additional annotator (who labeled every example). Then score each real annotator's quality using the GLAD method above.

\section{Why \methodabbrev{} can produce better estimates than other methods}
\begin{itemize}
    \item
    In settings with few (or only one) labels for an example, the agreement/Dawid-Skene/GLAD scores become unreliable \cite{paun2018comparing}. \methodabbrev{} can utilize additional information provided by a classifier that may be able to generalize to this example (especially if other dataset examples with similar feature values have more trustworthy consensus labels, e.g.\ if they received more annotations). \looseness=-1


    \item
    For examples that received a large number of annotations, CROWDLAB assigns less relative weight to the classifier predictions and its consensus quality score converges toward the observed annotator agreement. This quantity becomes more reliable when based on a large number of annotations \cite{paun2018comparing}, in which case relying on other sources of information becomes unnecessary. For examples where all annotations agree, an increase in the number of such annotations will typically correspond to an increased CROWDLAB consensus score. The \emph{Label Quality Score} alone fails to exhibit this desirable property.

    \item CROWDLAB uses weighted ensembling to combine annotations and classifier outputs. Countless prediction competitions have proven this to be among the most accurate/calibrated ways to combine different predictors.
    
    \item 
    Methods like Dawid-Skene estimate $K \times K$ confusion matrices per annotator, which may be statistically challenging when some annotators provide few labels \cite{paun2018comparing}. CROWDLAB merely estimates a single likelihood parameter $P$ shared across all classes/annotators in (\ref{eq:annotator_likelihood}) as well as a single per annotator  statistic $w_j$. Both can be better estimated from a limited number of observations. \looseness=-1

    \item 
    Generative-based methods like Dawid-Skene or GLAD are iterative algorithms, with high computational costs when their convergence is slow \cite{sinha2018fast, stephens2000dealing}. CROWDLAB does not require iterative updates and is  deterministic (for a given classifier).

\end{itemize}

\section{Experiments}
\label{sec:experiments}

\paragraph{Datasets. }
To evaluate various methods, we employ real-world multi-annotator data with naturally occurring label errors.
We run three benchmarks based on different subsets of the CIFAR-10H data \cite{cifar10h} which we call: \emph{Hardest}, \emph{Uniform}, \emph{Complete} (see Appendix 
\ref{sec:datasetdetails} for details and Appendix \ref{sec:uniformresults} and \ref{sec:completeresults} for additional results).
CIFAR-10H contains multiple labels for images in the CIFAR-10 test set \cite{cifar10}, obtained from a large set of new human annotators. 
As a source of ground truth labels, we simply use the corresponding labels for each image from the original CIFAR-10 dataset \cite{cifar10}. 
\citet{northcutt2021labelerrors} found the original CIFAR-10 labels to contain few errors in verification studies, and they have been adopted as ground truth labels in other research as well  \cite{labelerrordetection}.

\paragraph{Models. } 
To study how methods perform across different types of classifiers with varying accuracy, we applied every method twice, once using a ResNet-18 classifier \cite{he2016deep} and another time with a Swin Transformer model \cite{liu2021swin}. Both classifiers are trained on the same data (majority-vote consensus labels) in the same manner. 
Here the Swin Transformer represents a high quality model, whereas ResNet-18 represents a less accurate model (that is still commonly used in practice).

\paragraph{Metrics. }
To measure each of our three previously stated estimation tasks, we employ the following metrics: 
\begin{enumerate}
    \item 
    To evaluate \emph{how well methods can estimate consensus labels} from multiply-annotated data, we measure the \textbf{accuracy} of the inferred consensus label for each example against its ground truth label.
    \looseness=-1
    
    \item
    To evaluate \emph{how well methods can estimate the quality of each given consensus label}, we 
    compare estimated quality score $q_i$ for each example against a binary target indicating whether or not  the consensus label matches the ground truth label. If our goal is to use the quality scores to flag those examples whose consensus label is currently incorrect, this is a form of  information retrieval  \cite{labelerrordetection}. Thus our consensus quality scores are evaluated via precision/recall metrics: \textbf{AUROC},  \textbf{AUPRC}, and \textbf{Lift} at various cutoffs (which is directly proportional to Precision@$T$). To focus our evaluation purely on the estimation of label quality, throughout this section, we use each method to estimate quality scores for a single set of consensus labels established via majority vote. We always score the same of consensus labels here because our above evaluation already quantifies how good the consensus labels are from different methods, and we do not want this to confound our evaluation of how well different methods can estimate label quality. 
    
    \item
    To evaluate \emph{how well methods can estimate the quality of each annotator}, we measure the \textbf{Spearman correlation} between $a_j$ and $\text{ACC}_j$ over all annotators $j$, where: $a_j$ denotes our estimated annotator quality score (Sec. \ref{sec:aqs}) and $\text{ACC}_j$ denotes the accuracy of the $j$-th  annotator's chosen labels with respect to the ground truth labels (considering only the subset of examples labeled by annotator $j$). A method that achieves high Spearman correlation must produce annotator quality scores that are lower for those annotators whose labels tend to be wrong the most often.
\end{enumerate}

\section{Results}
\label{sec:results}

\begin{figure*}[tb]
\begin{subfigure}{.5\textwidth}
  \centering
  \includegraphics[width=\linewidth]{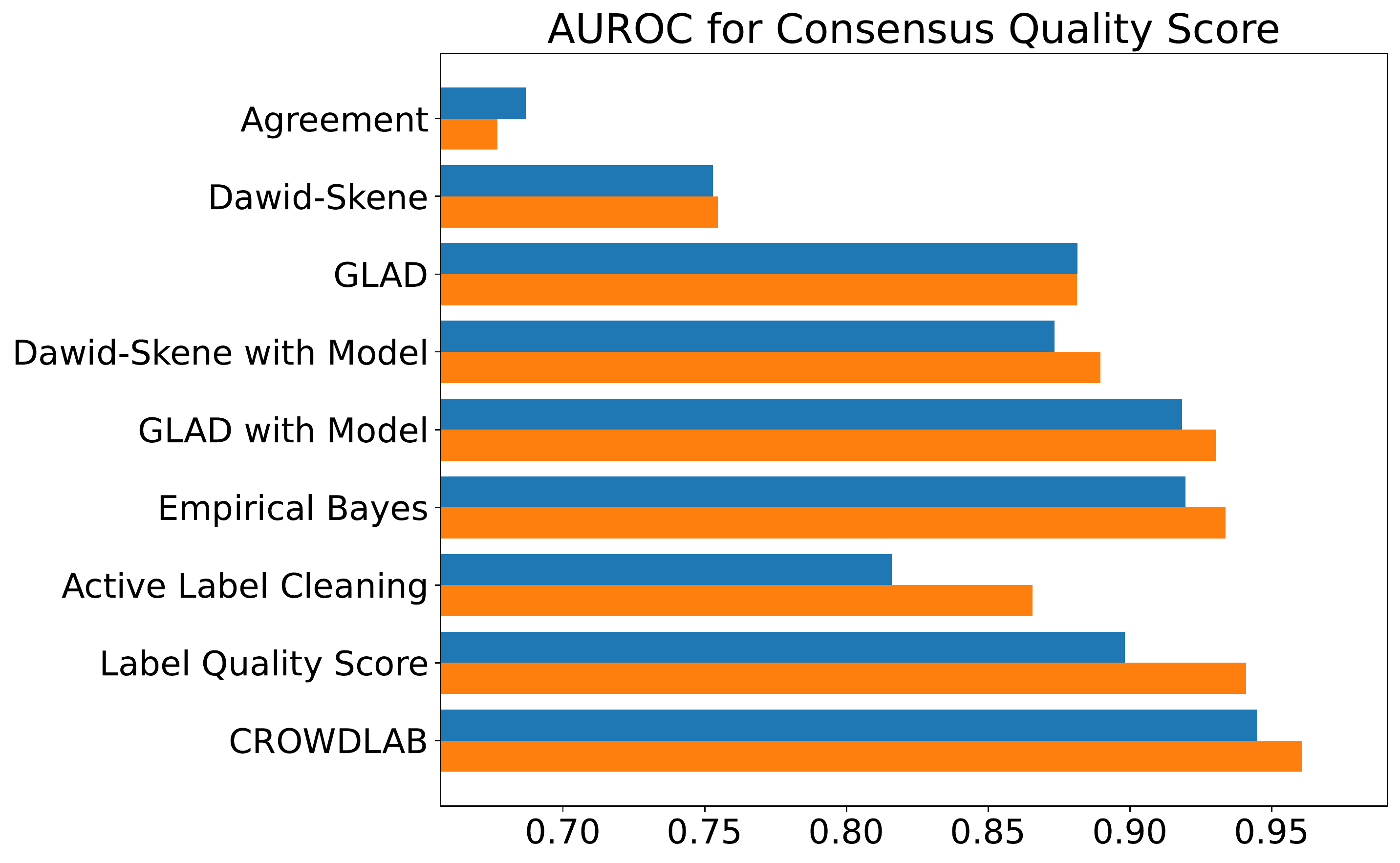} 
  \label{fig:auroc_worst}
    \vspace*{-0.7em}
\end{subfigure}
\begin{subfigure}{.5\textwidth}
  \centering
  \includegraphics[width=\linewidth]{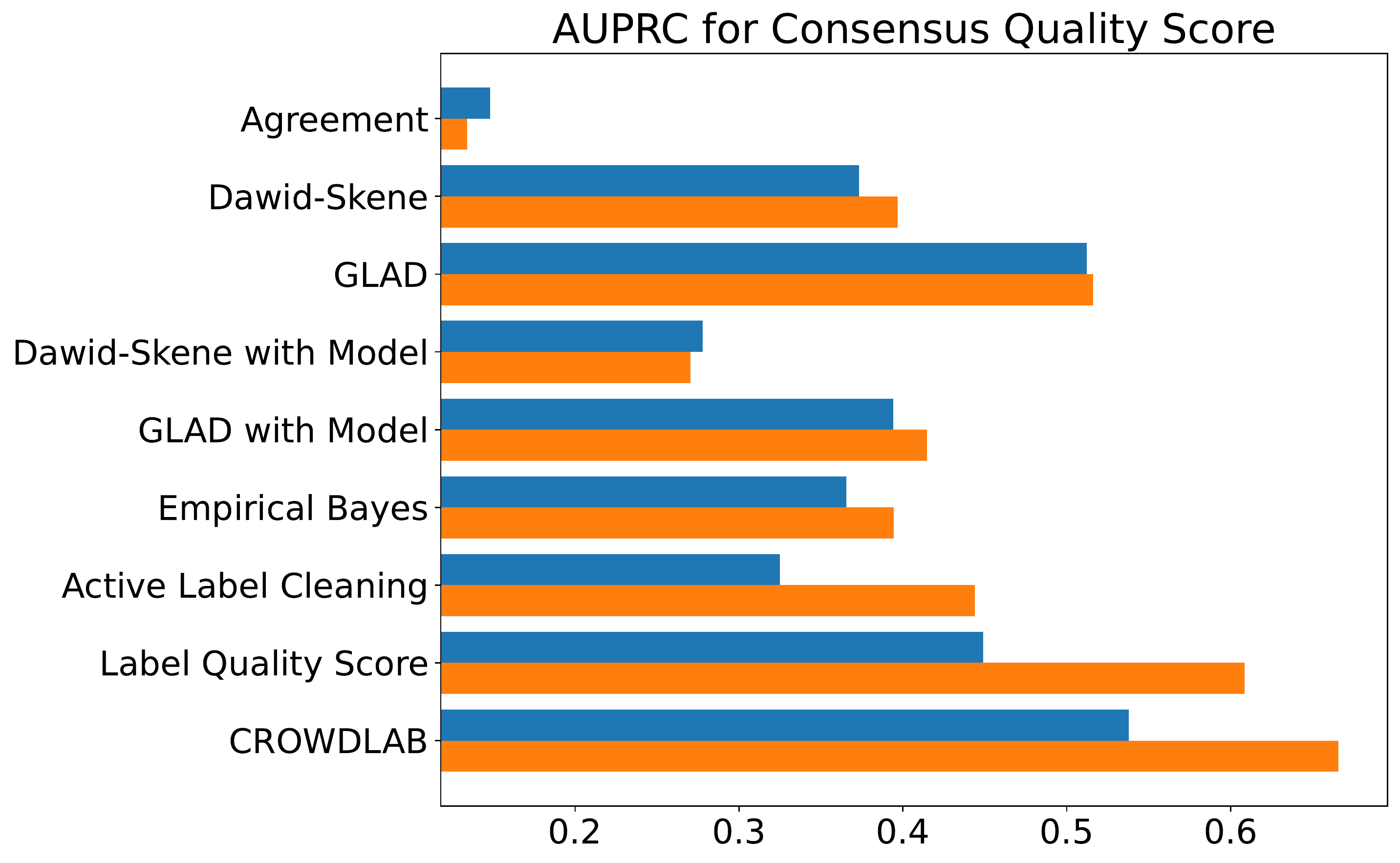} 
  \label{fig:auprc_worst}
        \vspace*{-0.7em}
\end{subfigure}
\begin{subfigure}{.5\textwidth}
  \centering
  \includegraphics[width=\linewidth]{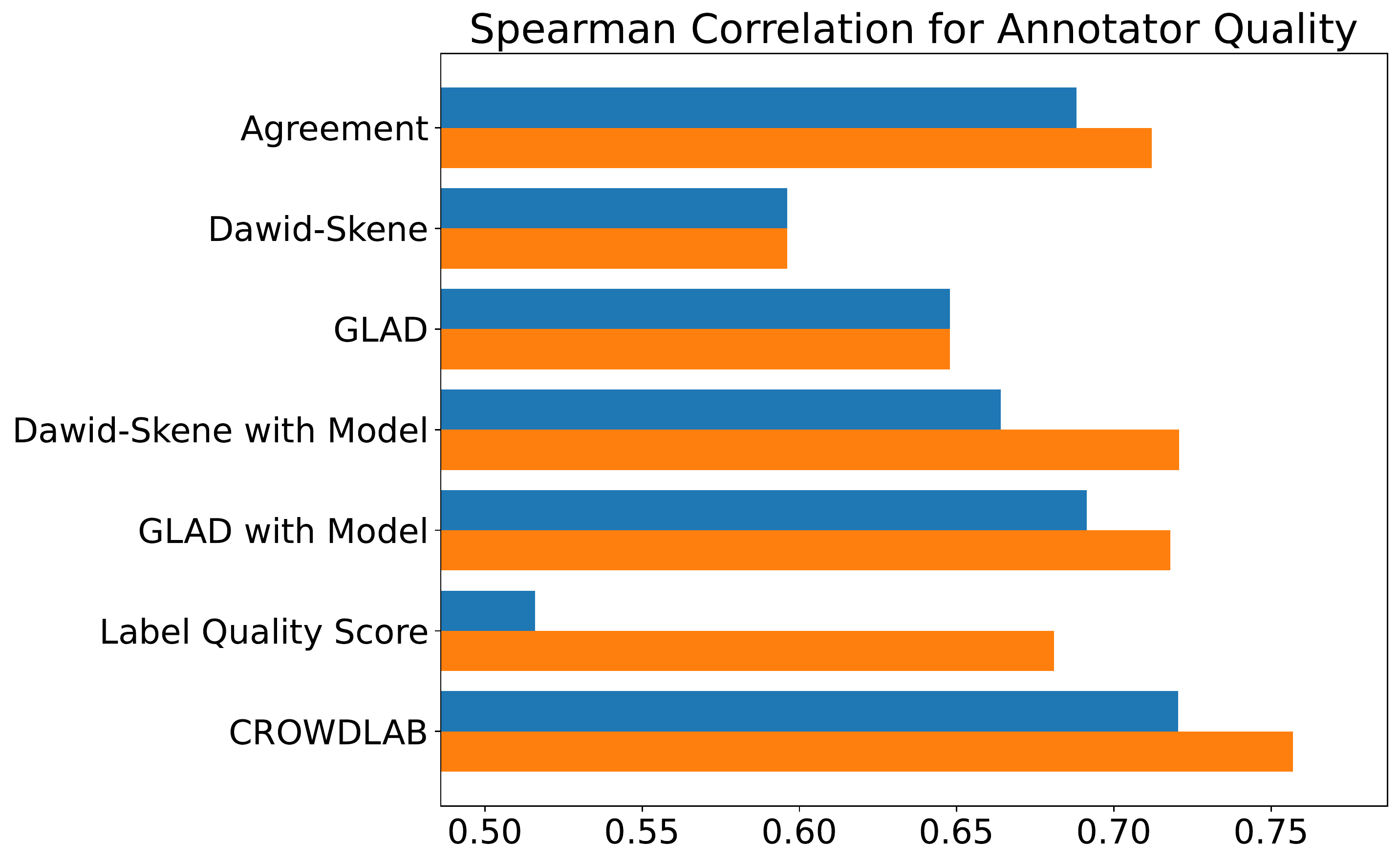}
  \label{fig:spearman_worst}
      \vspace*{-1.1em}
\end{subfigure} 
\begin{subfigure}{.5\textwidth}
  \centering
  \includegraphics[width=\linewidth]{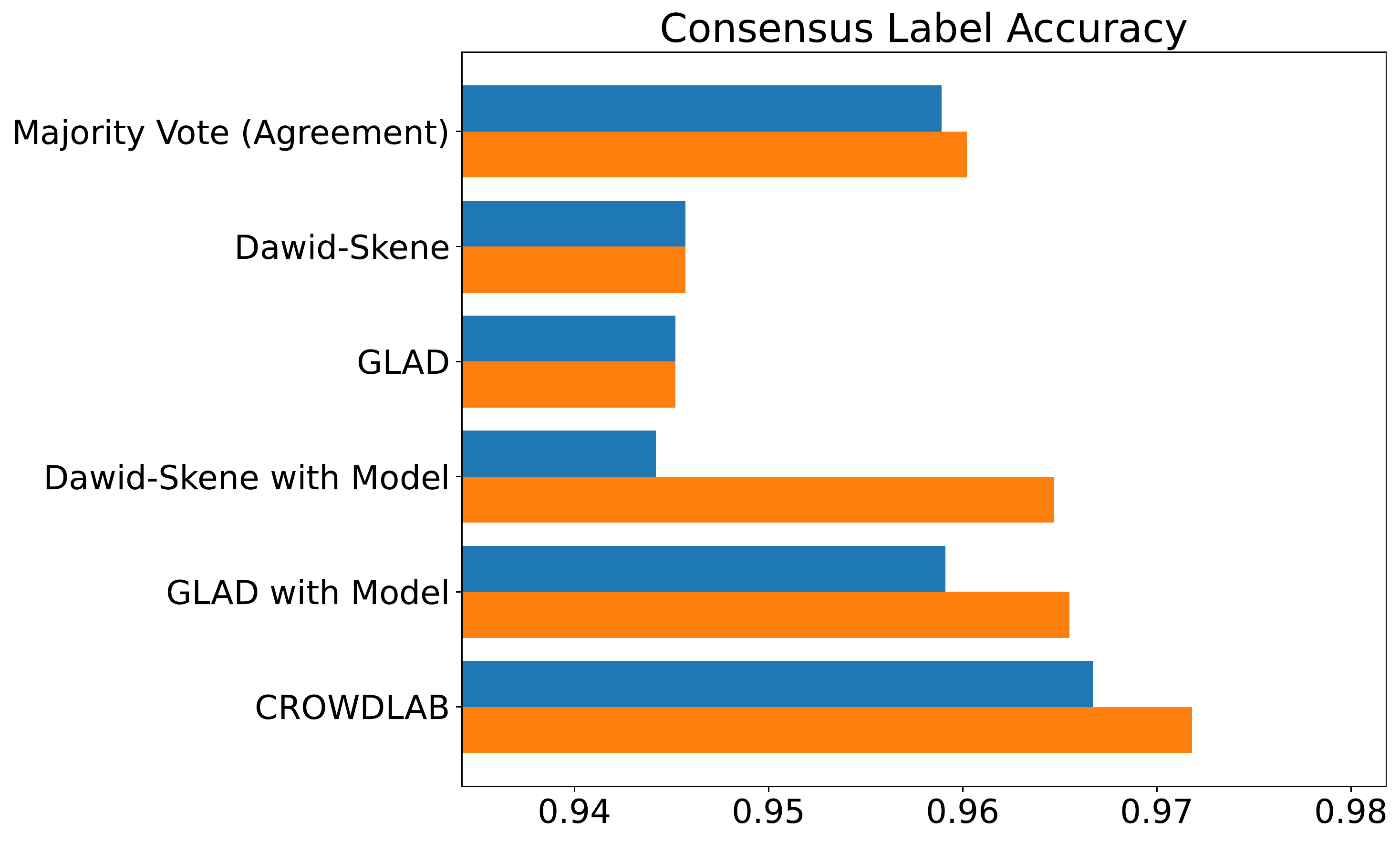}
  \label{fig:accuracy_worst}
    \vspace*{-1.1em}
\end{subfigure}
\begin{subfigure}{\textwidth}
  \centering
  \vspace*{-0em}
  \includegraphics[width=0.4\linewidth]{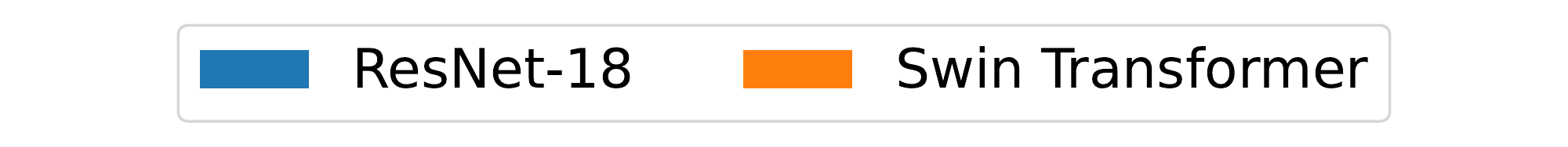} 
  \label{fig:legend_worst}
\end{subfigure}
\caption{Benchmarking methods to estimate consensus labels, their quality, and annotator quality 
on the \emph{Hardest} multi-annotator dataset.}
\label{fig:worst}
\end{figure*}

\begin{table*}[tb]
\centering 
\footnotesize
\begin{tabular}{l l c c c c c}
    \toprule 
    \bfseries Model & \bfseries Method & \bfseries Lift @ 10 & \bfseries Lift @ 50  & \bfseries Lift @ 100  & \bfseries Lift @ 300  & \bfseries Lift @ 500 \\
    \midrule
    \begin{filecontents*}{lift/worst_resnet_lift.csv}
dataset,model,quality_method,lift_at_10,lift_at_50,lift_at_100,lift_at_300,lift_at_500,lift_at_1000
worst_annotators,ResNet-18,Agreement,4.87,5.84,6.33,5.27,4.72,4.28
worst_annotators,ResNet-18,Dawid-Skene,12.89,11.79,13.26,10.74,8.51,5.14
worst_annotators,ResNet-18,GLAD,14.6,15.69,15.88,13.93,9.96,5.91
worst_annotators,ResNet-18,Dawid-Skene with Model,12.54,11.47,8.6,5.56,5.09,4.55
worst_annotators,ResNet-18,GLAD with Model,14.67,13.69,14.43,11.25,10.81,7.41
worst_annotators,ResNet-18,Empirical Bayes,12.17,12.17,11.68,11.11,10.27,7.3
worst_annotators,ResNet-18,Active Label Cleaning,17.03,19.95,16.3,10.22,7.88,5.16
worst_annotators,ResNet-18,Label Quality Score,19.46,21.41,19.22,13.38,10.22,6.5
worst_annotators,ResNet-18,CROWDLAB,24.33,22.38,17.76,14.27,11.82,7.76
\end{filecontents*}
\csvreader[head to column names]{lift/worst_resnet_lift.csv}{}
    { \model & \csvcoliii & \csvcoliv & \csvcolv & \csvcolvi & \csvcolvii & \csvcolviii \\} \\[-1em]
    \midrule 
    \begin{filecontents*}{lift/worst_swin_lift.csv}
dataset,model,quality_method,lift_at_10,lift_at_50,lift_at_100,lift_at_300,lift_at_500,lift_at_1000
worst_annotators,Swin,Agreement,2.51,6.03,6.28,4.86,4.17,4.1
worst_annotators,Swin,Dawid-Skene,12.89,14.36,14.55,10.99,8.66,5.25
worst_annotators,Swin,GLAD,14.6,15.69,15.88,14.11,10.04,5.89
worst_annotators,Swin,Dawid-Skene with Model,14.16,16.43,12.46,9.16,7.99,6.03
worst_annotators,Swin,GLAD with Model,8.7,17.39,17.1,15.36,11.71,8.17
worst_annotators,Swin,Empirical Bayes,12.56,9.55,11.06,11.81,11.76,7.94
worst_annotators,Swin,Active Label Cleaning,25.13,22.11,21.36,12.81,9.25,5.95
worst_annotators,Swin,Label Quality Score,25.13,22.61,21.86,16.92,12.81,7.96
worst_annotators,Swin,CROWDLAB,25.13,24.62,20.85,17.76,13.82,8.82
\end{filecontents*}
\csvreader[head to column names]{lift/worst_swin_lift.csv}{}
    { \model & \csvcoliii & \csvcoliv & \csvcolv & \csvcolvi & \csvcolvii & \csvcolviii \\} \\[-1em]
    \bottomrule \\
\end{tabular}
\caption{Evaluating the precision of various consensus quality scoring methods on the \emph{Hardest} dataset. Directly proportional to Precision@$T$, Lift@$T$ reports what fraction of the top-$T$ ranked consensus labels are actually incorrect,  normalized by the fraction of incorrect consensus labels expected for a random set of examples.}
\label{tab:lift}
\end{table*}

Figures \ref{fig:worst}, \ref{fig:uniform}, \ref{fig:complete}, and Tables \ref{tab:lift}, \ref{tab:lift_uniform},  \ref{tab:lift_complete} demonstrate that CROWDLAB overall performs the best across our evaluations for consensus and annotator quality scores, and also typically produces the most accurate consensus labels. 
For most methods considered in this paper, all evaluation metrics improve when used with the Swin Transformer vs.\ ResNet-18 model. This illustrates how a better classifier can be utilized to get more improvement in consensus labels and consensus/annotator quality estimates. Effective methods for multi-annotator analysis must remain compatible with future innovations in classifier technology. 

Considering only classifier predictions and consensus labels  (rather than individual annotator information), the \emph{Label Quality Score} also effectively estimates consensus quality when we have an accurate model (Swin Transformer). 
Predictions from a strong classifier suffice to estimate label quality without additional information provided by individual annotators \cite{labelerrordetection}. 
However \emph{Label Quality Score} performs worse than other methods with a lower accuracy classifier  (ResNet-18). This demonstrates the value of accounting for the individual annotations and overall model accuracy in CROWDLAB, which performs well relative to other methods regardless of the classifier's accuracy. 
Treating the classifier as an additional annotator for the Dawid-Skene and GLAD methods improves their performance, but not enough to match CROWDLAB, which better accounts for the classifier's confidence.  While the \emph{Empirical Bayes} method also accounts for classifier confidence to augment Dawid-Skene, it similarly unable to match CROWDLAB, demonstrating  why our method considers \emph{how much} to weigh the model based on its estimated trustworthiness relative to the annotators.  

In Appendix \ref{sec:variants}, we also compare CROWDLAB against a variant of our method which lacks the per-annotator quality estimation (i.e.\ all annotator weights $w_j$ are equal). Empirically, this  variant underperforms as it estimates \emph{too little} information about the annotators. 
On another \emph{Uniform} dataset in which there are 1-5 annotations for each example occurring with equal frequencies, CROWDLAB is able to produce better estimates for tasks (1)-(3) than the other methods considered here (results in Appendix \ref{sec:uniformresults}). 
On another \emph{Complete} dataset with many more ($\sim 50$) annotations per example, such that simple annotator agreement and majority vote produce  highly accurate estimates, CROWDLAB retains its strong performance compared to other methods (results in Appendix \ref{sec:completeresults}).
In Appendix \ref{sec:truelabels}, we run all methods with an unrealistically accurate classifier on all datasets. This setting favors the \emph{Label Quality Score}, but we find that CROWDLAB still outperforms the other methods. 
This breadth of settings highlights the utility of CROWDLAB across a wide range of applications involving fair/stellar classfier models and varying numbers of data annotations.


\section{Discussion}

Unlike other ways to utilize classifiers with crowdsourcing algorithms, CROWDLAB considers a model's estimated confidence and how accurate it is relative to individual annotators. Methods such as Dawid-Skene (or GLAD) with Model account for the model predictions, but fail adjust for model confidence and accuracy which is properly done by CROWDLAB.
Our proposed methodology is compatible with any classifier and training strategy, ensuring its out-of-the-box performance will improve as new models and training tricks are invented. This is vital as classifier technology continues to improve, and ensures CROWDLAB can be applied to diverse data (image, text, tabular, audio, etc).

The efficacy of CROWDLAB depends on being able to train a performant classifier, unlike generative models of annotator statistics. 
Fortunately, training good classifiers is easy with modern AutoML \cite{agtabular} and techniques for calibration, data augmentation, and transfer learning \cite{thulasidasan2019mixup}. 
As with most classification projects, CROWDLAB users should remain wary of overconfident model predictions with limited ability to generalize, which may lead to overly optimistic estimates of quality. Ensuring predictions are out-of-sample  helps mitigate this. \looseness=-1


\FloatBarrier
\newpage
\clearpage
\newpage
\bibliographystyle{abbrvnat}
\bibliography{multiannotator}
\balance{}

\clearpage \newpage
\beginsupplement
\onecolumn
\appendix

\def\toptitlebar{\hrule height1pt \vskip .2in} 
\def\bottomtitlebar{\vskip .22in \hrule height1pt \vskip .3in} 

\thispagestyle{plain}

\setcounter{page}{1}
\pagenumbering{arabic}
\setlength{\footskip}{20pt}  
\begin{center}
\toptitlebar
{\Large \bf \appendixtitle 
}
\bottomtitlebar
\end{center}

\FloatBarrier

\section{Experiment Details}
\label{sec:details}

Our experiments employ two of the most currently popular architectures for image classification, which are intended to be representative of different types of models one might use in practice.   
Training of the Swin Transformer and ResNet classifiers was done as by \citet{labelerrordetection}, using 5-fold cross-validation starting with ImageNet-pretrained weights fine-tuned in each fold via AutoML \citep{agtabular}. 
When establishing consensus labels via majority vote, we break ties for an example by favoring the class which was annotated more often overall across the dataset.
We do not evaluate annotator quality scores from the \emph{Active Label Cleaning} method because rating annotators was left as future work in the paper of \citet{bernhardt2022active}. Annotator quality estimates from the \emph{Empirical Bayes} approach match those from \emph{Dawid-Skene} and are also omitted from our plots.

Note that all metrics discussed in Section \ref{sec:experiments} are for evaluation purposes only, and would not be computable in real applications of our methodology due to a lack of ground truth labels. 
For evaluating consensus quality scores, AUROC measures how well these scores are able to differentiate correct and incorrect consensus labels. AUPRC accounts for the precision/recall of the consensus quality scores in flagging an incorrect consensus label, in a manner that is more sensitive to proportion of errors in the majority-vote consensus label errors than AUROC \cite{davis2006relationship}. The Lift at $T$ metric measures how much more likely we are to encounter an incorrect consensus label among the top $T$ ranked examples that have the worst consensus quality score.
Although our evaluation of consensus quality estimation  is applied to majority-vote consensus labels in Section \ref{sec:experiments}, each method could be used to estimate the quality of consensus labels derived via another approach.

\subsection{Datasets} 
\label{sec:datasetdetails}

The original CIFAR-10 dataset \cite{cifar10} is fairly easy to label \cite{northcutt2021labelerrors}. Thus annotator agreement on the complete CIFAR-10H data \cite{cifar10h} is unrealistically high for a representative multi-annotator benchmark. The images are not only easy to label, but there is also an uncommonly large number of annotators ($\sim 50$) per image in CIFAR-10H. Labeling budgets are typically too small to have so many annotators review each example.
Hence our primary benchmark uses a subset of the CIFAR-10H annotations. This subset starts with the 25 worst annotators and then incrementally add annotators from worst to best (based on their accuracy vs.\ ground-truth labels) until each of the 10,000 examples have at least 1 annotation (resulting in a dataset with 511 annotators in total). During this process, we restricted the selection of each new annotator to add to the current subset to only those which labeled at least one example also labeled by one of the annotators in the current subset. 
We call this variation of CIFAR-10H the \emph{Hardest} dataset benchmarked in this paper, and believe it is more representative of real-world data labeling applications, where the proportion of label errors tends to be far higher than in CIFAR-10H and the number of annotators far lower \cite{hivemind}. 

To ensure the robustness of our conclusions, we also evaluated all methods on two other datasets: a \emph{Uniform} subset of CIFAR-10H (only considering some randomly chosen annotators such that each example has between 1-5 annotations with an equal number of examples receiving 1 annotation, 2 annotations, etc.), and the \emph{complete} CIFAR-10H dataset (with all annotator labels, which is far more than typically collected in most applications). Results for these other datasets are in Appendix \ref{sec:uniformresults} and \ref{sec:completeresults}, and are based on separate classifier models trained for each dataset. In all cases, we only consider images from the \emph{test set} of CIFAR-10 (here treated as multiply-labeled training data), since these are the only images labeled by many annotators in CIFAR-10H. \looseness=-1


\begin{table}[H]
\centering 
\begin{tabular}{l c}
    \toprule 
\bfseries Labels predicted by & \bfseries Accuracy (w.r.t. ground truth labels) \\
    \midrule 
ResNet-18 & 0.879 \\
Swin Transformer & 0.940 \\
Swin Transformer trained with true labels & 0.948 \\
Annotator (Average) & 0.909 \\
    \bottomrule \\
\end{tabular}
\caption{Classification accuracy for the \emph{Hardest} dataset achieved by various predictors: ResNet-18 and Swin Transformer classifiers trained on majority-vote consensus labels (i.e.\ the models used in the benchmark results of Figure \ref{fig:worst}), Swin Transformer trained on true labels, which represents an unrealistically good classifier (see Appendix \ref{sec:truelabels}), as well as the average annotator in the dataset.}
\label{tab:modelaccuracy}
\end{table}
\vspace*{8mm}


\section{Results for Uniform Dataset}
\label{sec:uniformresults}

To evaluate our methods in another setting, we construct a different subset of CIFAR10-H and re-run our benchmark on this new dataset. In this \emph{Uniform} dataset, each example now has between 1 to 5 labels, where the number of labels per example are uniformly distributed. This dataset contains 421 annotators and 10,000 examples. Here the annotators are just randomly selected from the CIFAR10-H pool, and are thus higher quality than in the \emph{Hardest} dataset.
The following results demonstrate that CROWDLAB is also the best method overall for this  \emph{Uniform} dataset. 

\vspace*{8mm}

\begin{figure}[H]
\begin{subfigure}{.5\textwidth}
  \centering
  \includegraphics[width=\linewidth]{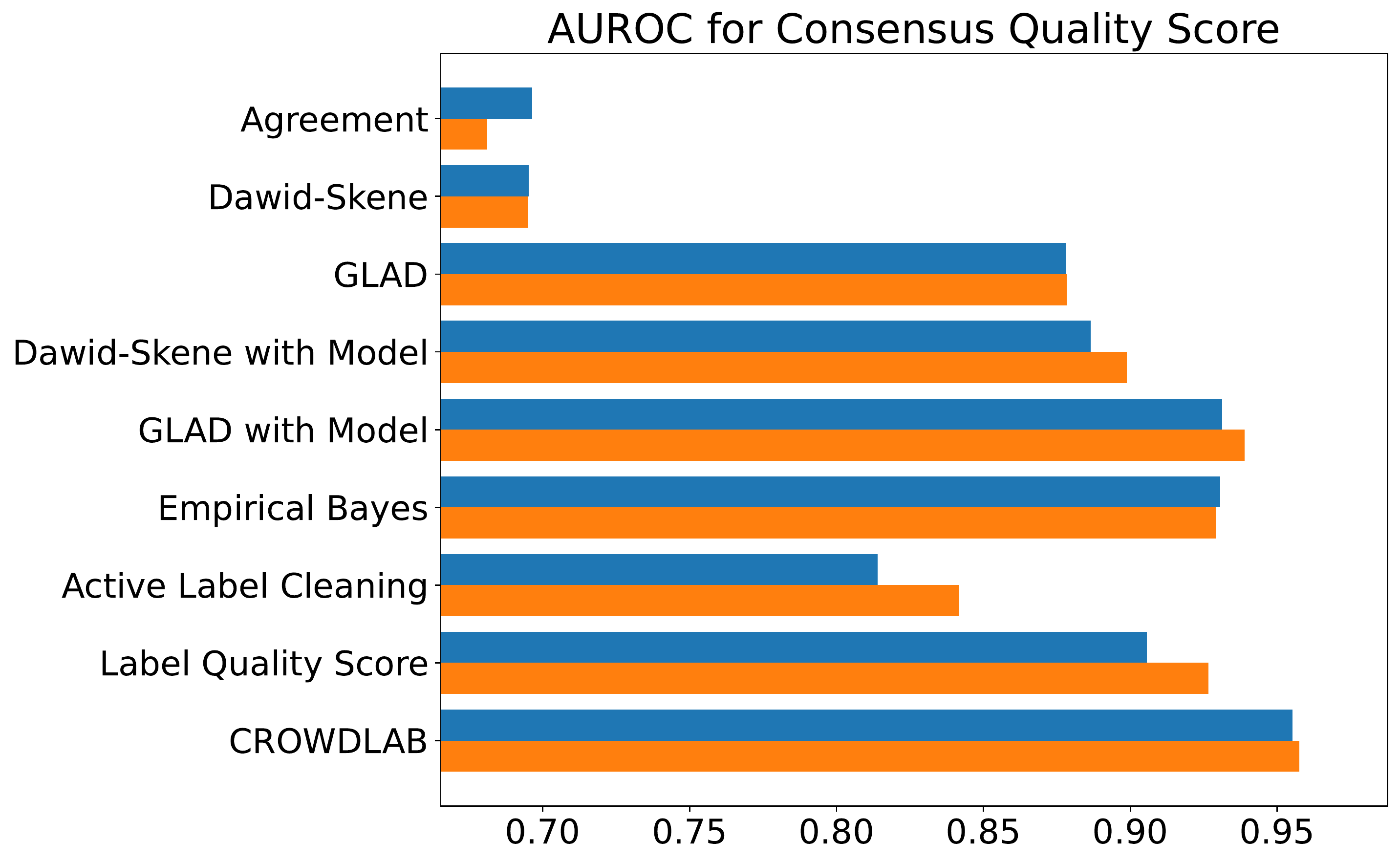} 
  \label{fig:auroc_uniform}
\end{subfigure}
\begin{subfigure}{.5\textwidth}
  \centering
  \includegraphics[width=\linewidth]{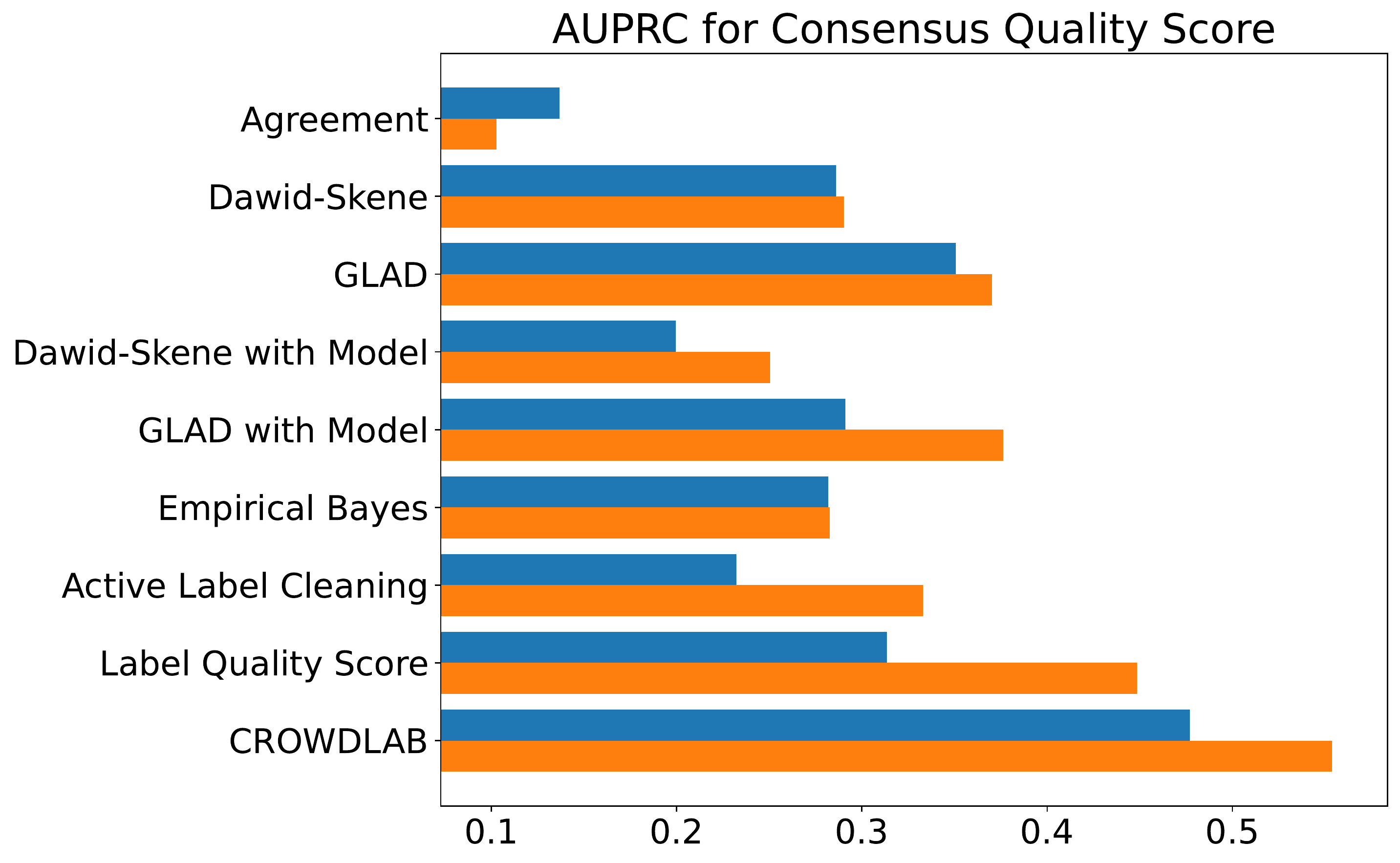} 
  \label{fig:auprc_uniform}
\end{subfigure}
\vspace*{-2mm}
\begin{subfigure}{.5\textwidth}
  \centering
  \includegraphics[width=\linewidth]{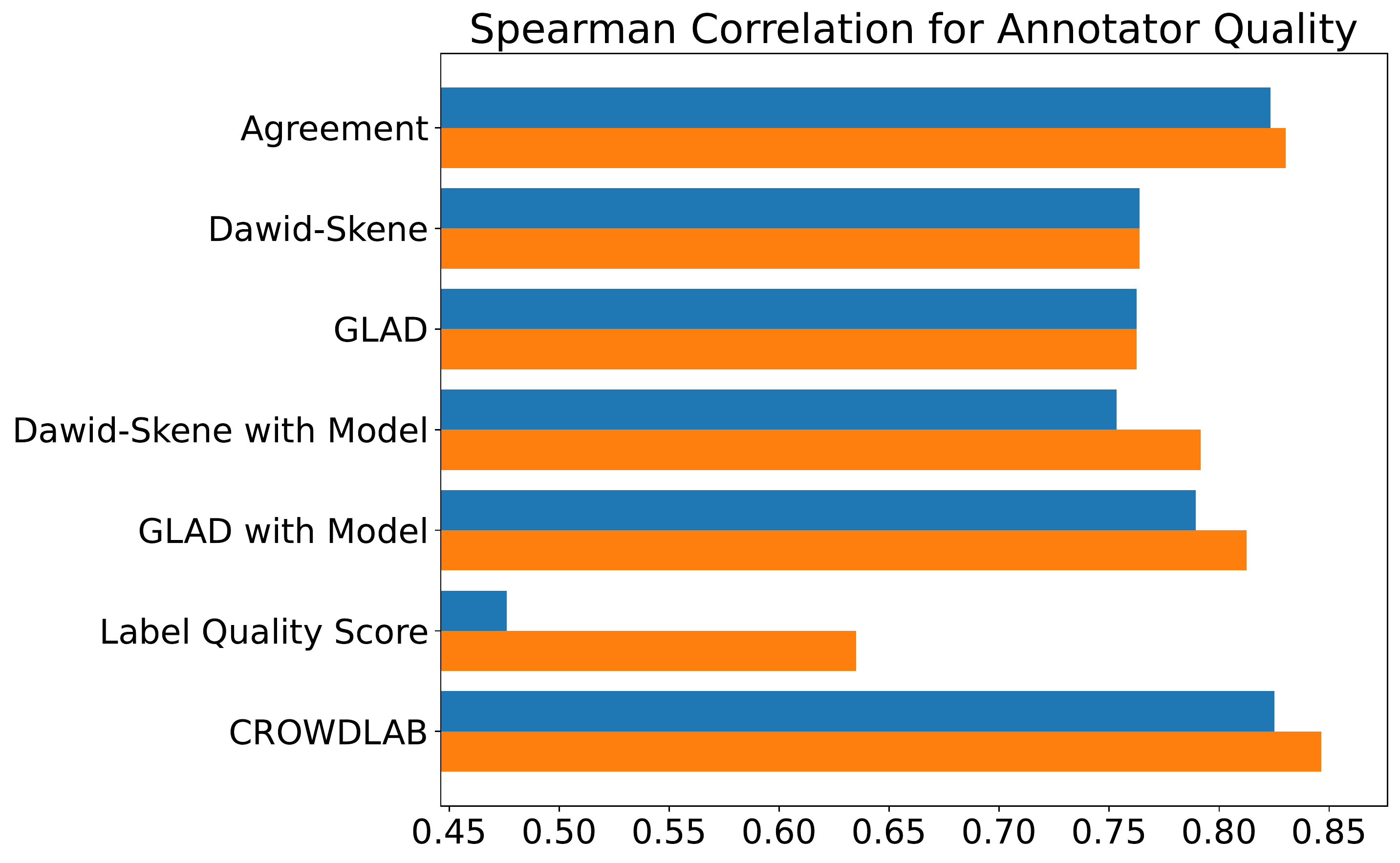}
  \label{fig:spearman_uniform}
\end{subfigure} 
\begin{subfigure}{.5\textwidth}
  \centering
  \includegraphics[width=\linewidth]{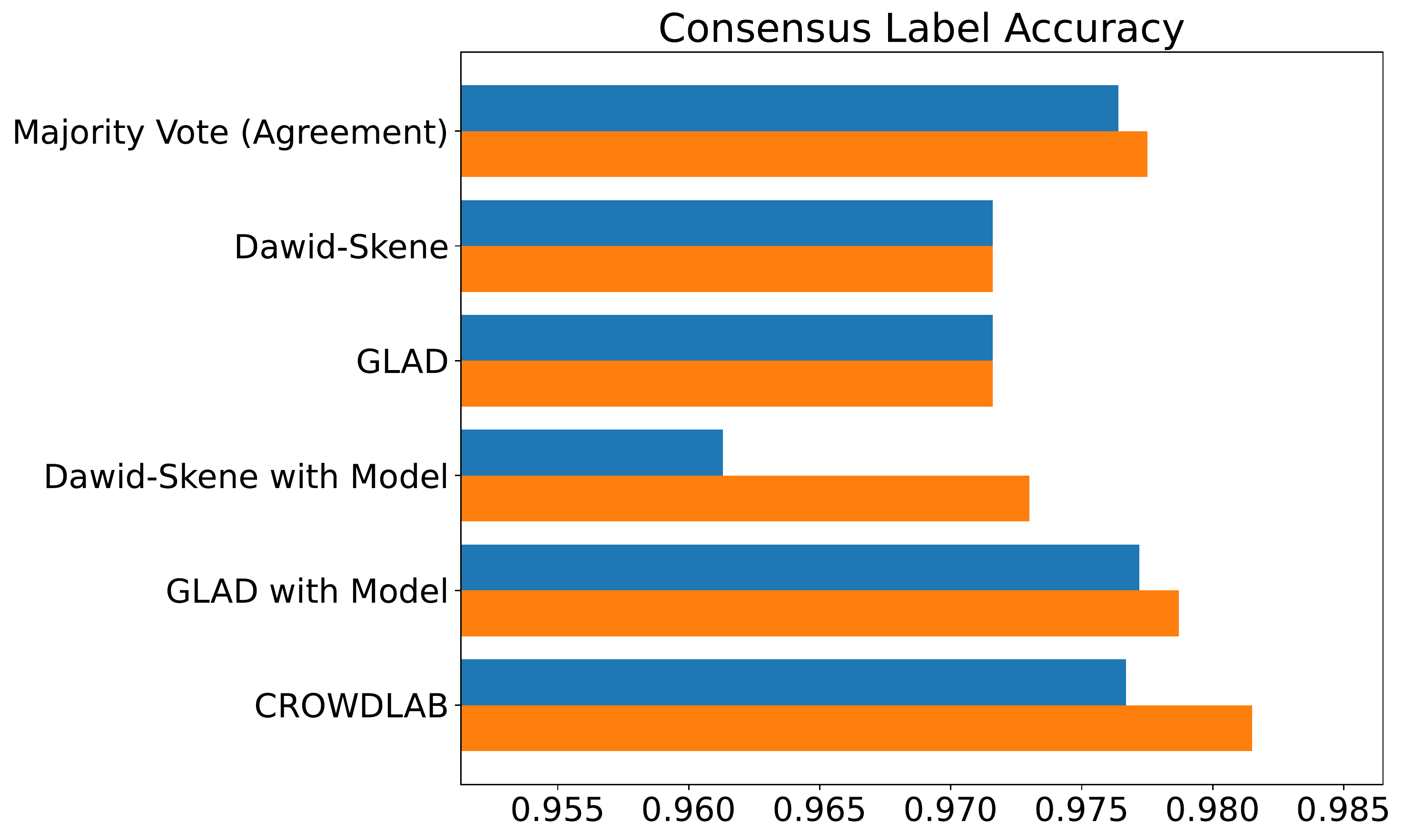}
  \label{fig:accuracy_uniform}
\end{subfigure}
\begin{subfigure}{\textwidth}
\vspace*{-1mm}
  \centering
  \includegraphics[width=0.4\linewidth]{figures/legend_wide.pdf} 
  \label{fig:legend_uniform}
\end{subfigure}
\vspace*{-5mm}
\caption{Benchmarking methods to estimate consensus labels, their quality, and annotator quality 
on the \emph{Uniform} dataset.}
\label{fig:uniform}
\end{figure}

\clearpage 

\begin{table}[H]
\centering 
\footnotesize
\begin{tabular}{l l c c c c c}
    \toprule 
    \bfseries Model & \bfseries Quality Method & \bfseries Lift @ 10 & \bfseries Lift @ 50  & \bfseries Lift @ 100  & \bfseries Lift @ 300  & \bfseries Lift @ 500 \\
        \midrule
    \begin{filecontents*}{lift/uniform_resnet_lift.csv}
dataset,model,quality_method,lift_at_10,lift_at_50,lift_at_100,lift_at_300,lift_at_500,lift_at_1000
uniform_1_5,ResNet-18,Agreement,21.19,11.86,9.75,9.04,7.12,4.62
uniform_1_5,ResNet-18,Dawid-Skene,31.69,24.65,19.01,12.09,8.52,5.46
uniform_1_5,ResNet-18,GLAD,31.69,22.54,25.7,13.03,9.23,5.32
uniform_1_5,ResNet-18,Dawid-Skene with Model,10.34,11.37,7.24,5.43,4.86,3.77
uniform_1_5,ResNet-18,GLAD with Model,17.54,18.42,15.79,13.01,12.72,8.33
uniform_1_5,ResNet-18,Empirical Bayes,12.71,20.34,18.22,13.98,11.44,7.46
uniform_1_5,ResNet-18,Active Label Cleaning,33.9,24.58,16.1,10.03,7.88,5.59
uniform_1_5,ResNet-18,Label Quality Score,33.9,26.27,22.46,12.43,9.75,6.78
uniform_1_5,ResNet-18,CROWDLAB,42.37,33.05,27.97,16.95,13.81,8.64
\end{filecontents*}
\csvreader[head to column names]{lift/uniform_resnet_lift.csv}{}
    {  \model & \csvcoliii & \csvcoliv & \csvcolv & \csvcolvi & \csvcolvii & \csvcolviii \\} \\[-1em]
        \midrule 
        \begin{filecontents*}{lift/uniform_swin_lift.csv}
dataset,model,quality_method,lift_at_10,lift_at_50,lift_at_100,lift_at_300,lift_at_500,lift_at_1000
uniform_1_5,Swin,Agreement,8.89,8.0,7.56,7.85,6.49,4.36
uniform_1_5,Swin,Dawid-Skene,28.17,27.46,20.07,11.74,8.45,5.49
uniform_1_5,Swin,GLAD,35.21,25.35,27.11,13.03,9.23,5.32
uniform_1_5,Swin,Dawid-Skene with Model,33.33,16.3,12.22,8.89,8.15,6.93
uniform_1_5,Swin,GLAD with Model,23.47,23.47,23.0,19.87,14.74,8.4
uniform_1_5,Swin,Empirical Bayes,13.33,17.78,17.78,14.96,12.44,7.87
uniform_1_5,Swin,Active Label Cleaning,44.44,30.22,24.0,14.07,10.13,6.62
uniform_1_5,Swin,Label Quality Score,35.56,32.89,29.33,18.67,13.6,8.0
uniform_1_5,Swin,CROWDLAB,40.0,35.56,32.0,21.19,15.2,8.84
\end{filecontents*}
\csvreader[head to column names]{lift/uniform_swin_lift.csv}{} 
           { \model & \csvcoliii & \csvcoliv & \csvcolv & \csvcolvi & \csvcolvii & \csvcolviii \\}
 \\[-1em]
    \bottomrule \\[-0.5em]
\end{tabular}
 \caption{Evaluating the precision of various consensus quality scoring methods on the \emph{Uniform} dataset. Lift@$T$ is directly proportional to Precision@$T$, and reports what fraction of the top-$T$ ranked consensus labels are actually incorrect normalized by the fraction of incorrect consensus labels expected for a random set of examples.}
  \label{tab:lift_uniform}
\end{table}

\newpage
\section{Results for Complete Dataset}
\label{sec:completeresults}

We also evaluate our methods on the full original CIFAR-10H dataset \cite{cifar10h}. This \emph{Complete} dataset contains 2571 annotators where each annotator labels 200 examples, such that each of the 10,000 images has approximately 50 annotations. The \emph{Complete} dataset has by far the highest number of annotations per example out of all the datasets considered in this paper. Far less annotations per example are available in most real-world multi-annotator datasets due to limited labeling budgets.

With so many annotations per example, basic annotator agreement methods are highly effective. CROWDLAB works similarly well, highlighting its adaptive nature across different datasets with few vs. many annotations per example. Even in this \emph{Complete} dataset where majority-vote consensus labels should be extremely accurate, CROWDLAB consensus labels are even more accurate (regardless whether a Swin Transformer or  ResNet-18 model is used to augment the annotators).

Even though CROWDLAB is straightforward and only estimates one  parameter per annotator ($w_j$) and two others in total ($P$ and $w_{\mathcal{M}}$), it performs well on this annotation-rich dataset. This indicates CROWDLAB is not too simple, given that methods which estimate many more parameters like Dawid-Skene and GLAD do not perform better on this \emph{Complete} dataset, even though there is no shortage of annotations to learn from. 
CROWDLAB's fewer number of parameters entail a key advantage for most datasets which have fewer annotations, and do not cause it to lag behind richer generative methods like Dawid-Skene/GLAD in this setting with unusually many annotations. 

\vspace*{8mm}

\begin{figure}[H]
\begin{subfigure}{.5\textwidth}
\hspace*{0.1mm}
  \centering
  \includegraphics[width=\linewidth]{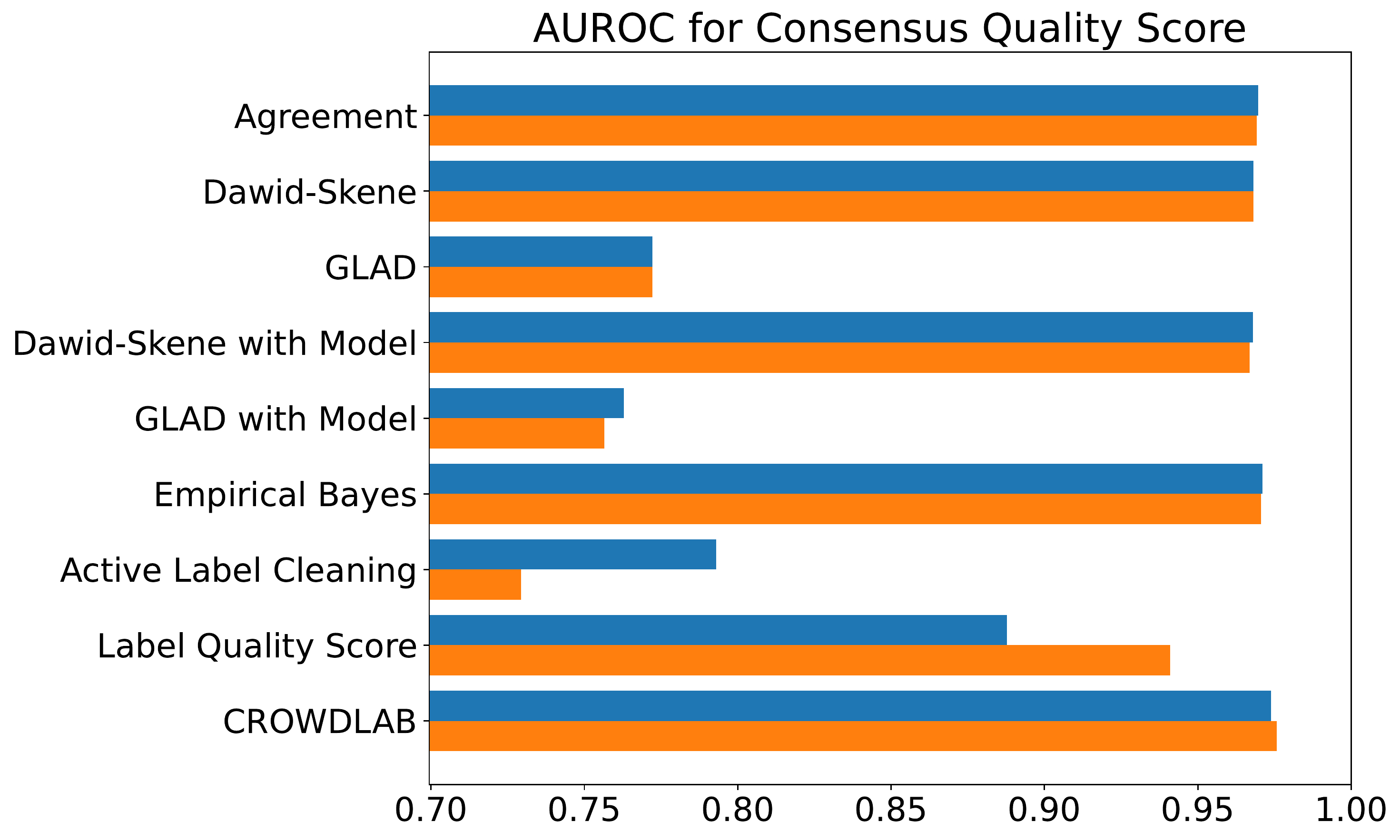} 
  \label{fig:auroc_complete}
\end{subfigure}
\begin{subfigure}{.5\textwidth}
\hspace*{1mm}
  \centering
  \includegraphics[width=\linewidth]{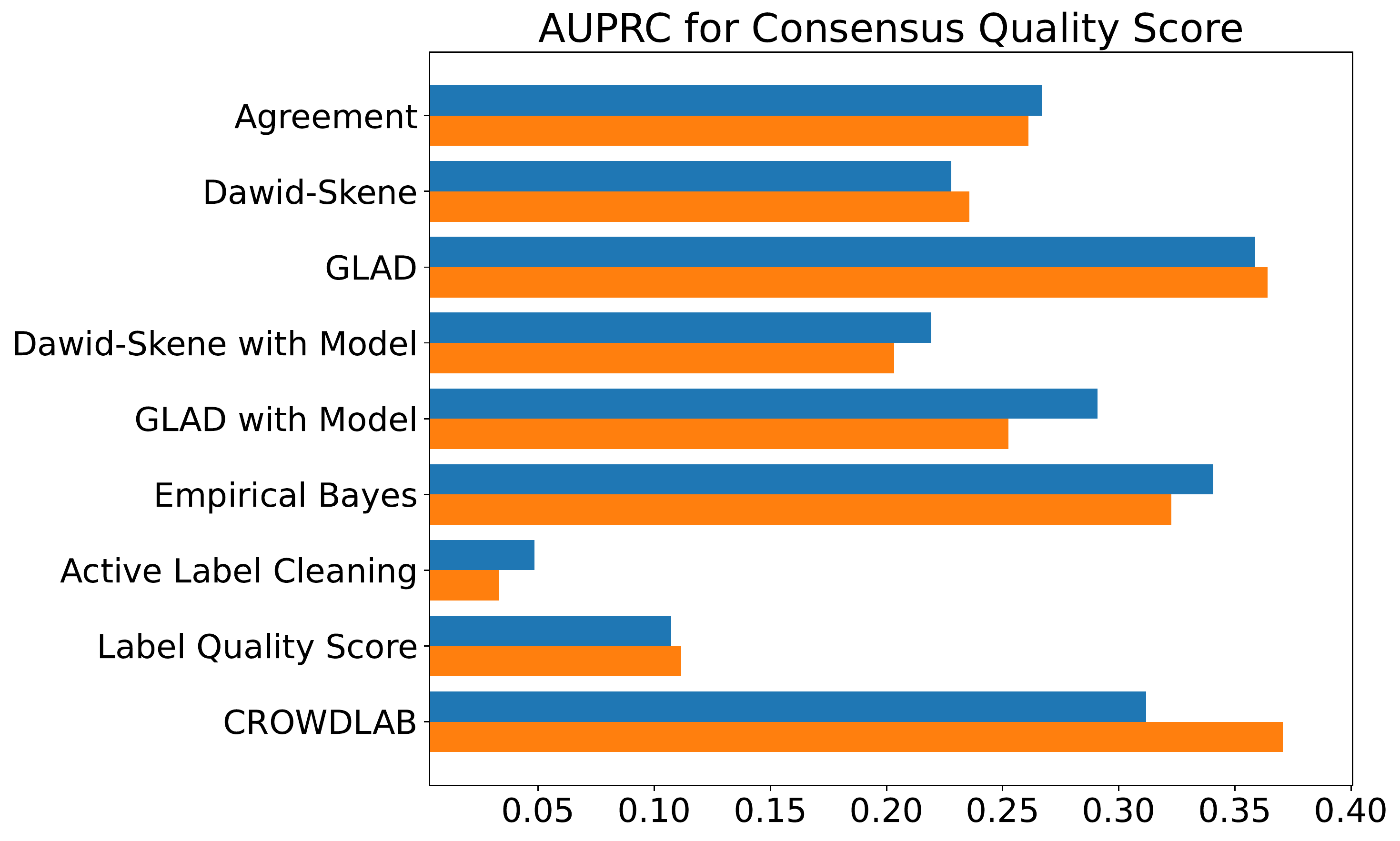} 
  \label{fig:auprc_complete}
\end{subfigure}
\vspace*{-2mm}
\begin{subfigure}{.5\textwidth}
  \centering
  \includegraphics[width=\linewidth]{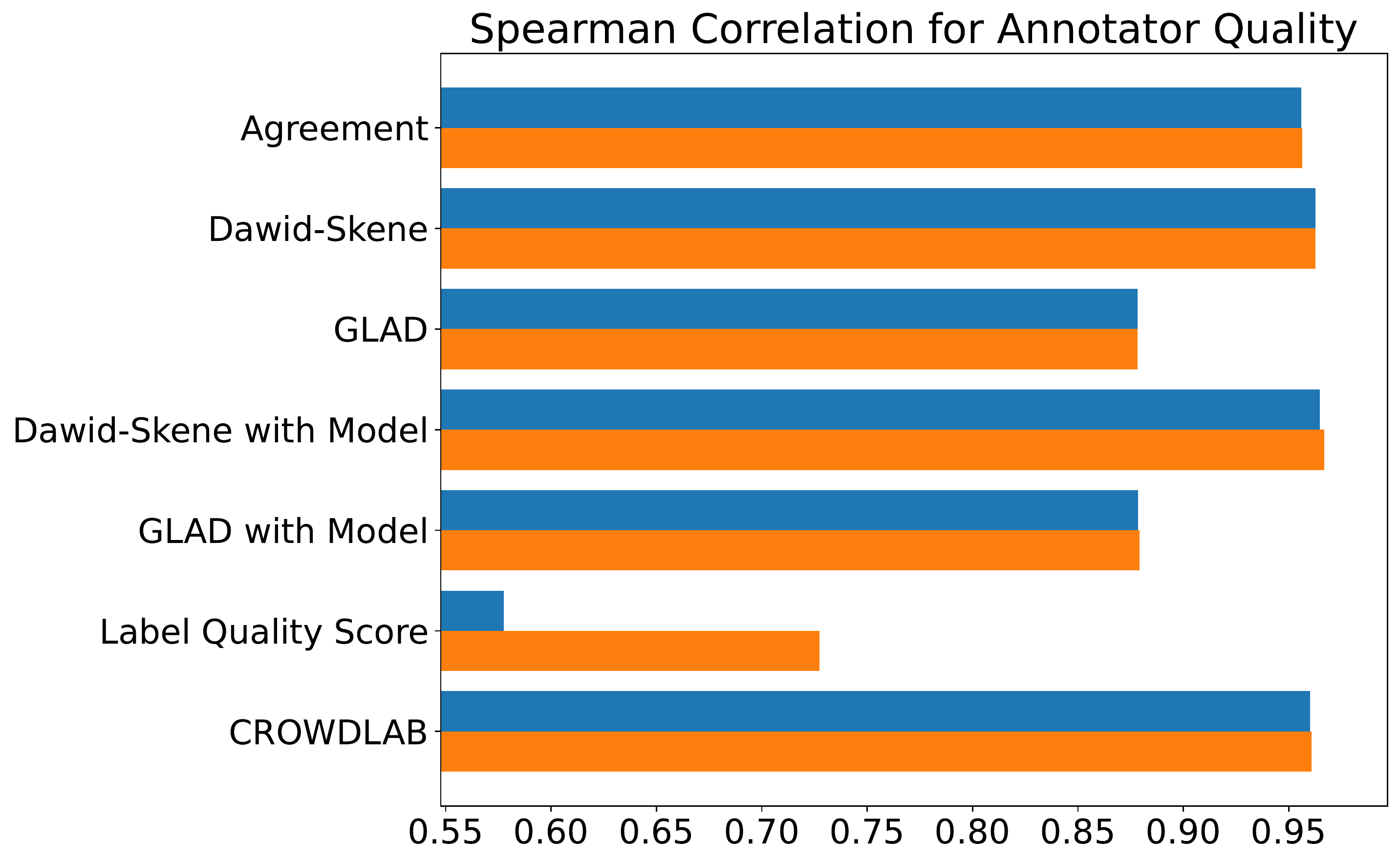}
  \label{fig:spearman_complete}
\end{subfigure} 
\begin{subfigure}{.5\textwidth}
\vspace*{-1mm}
  \centering
  \includegraphics[width=\linewidth]{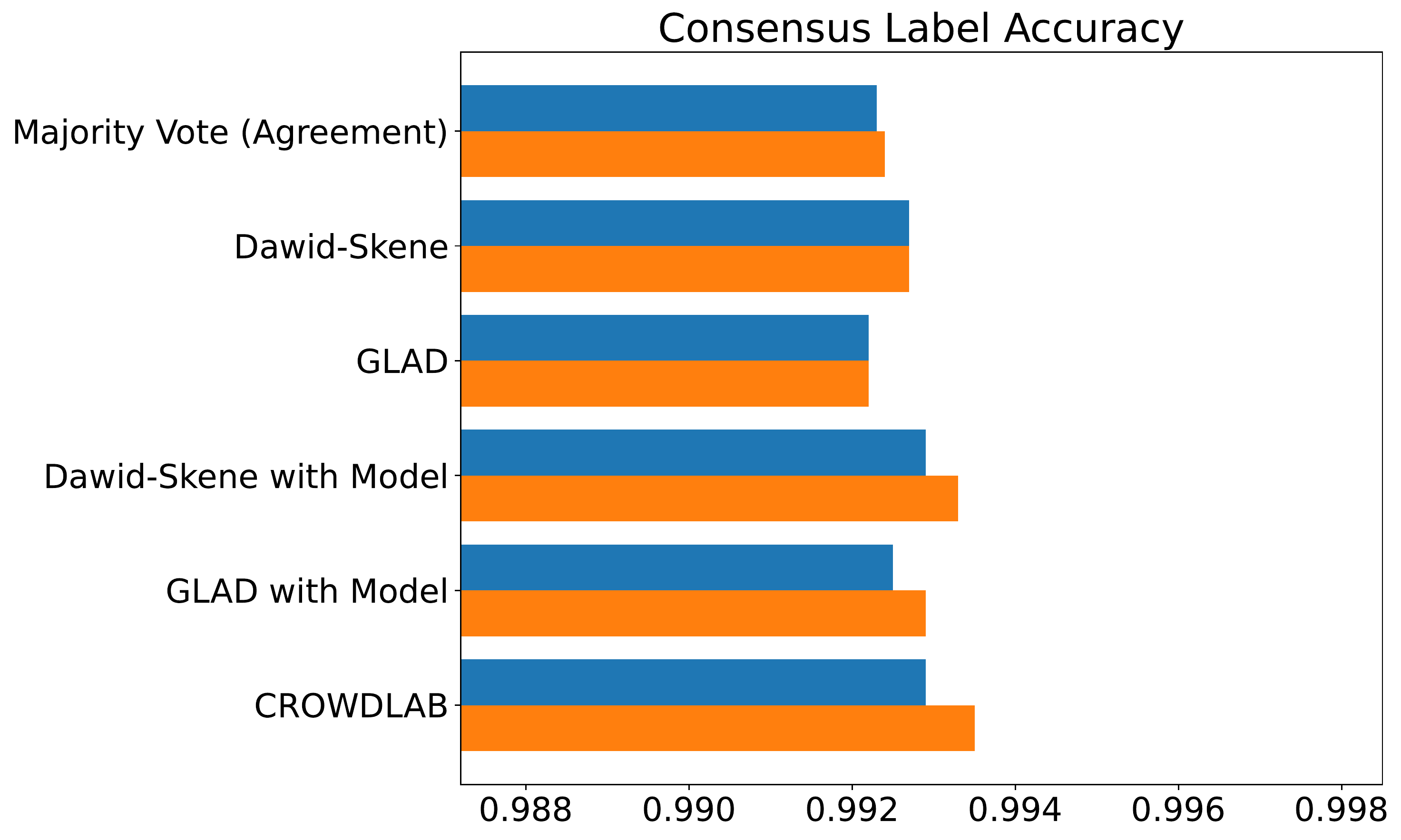}
  \label{fig:accuracy_complete}
\end{subfigure}
\begin{subfigure}{\textwidth}
  \centering
  \includegraphics[width=0.4\linewidth]{figures/legend_wide.pdf} 
  \label{fig:legend_complete}
\end{subfigure}
\vspace*{-5mm}
\caption{Benchmarking methods to estimate consensus labels, their quality, and annotator quality 
on the \emph{Complete} dataset.}
\label{fig:complete}
\end{figure}

\clearpage

\begin{table}[H]
\centering 
\footnotesize
\begin{tabular}{l l c c c c c}
    \toprule 
    \bfseries Model & \bfseries Quality Method & \bfseries Lift @ 10 & \bfseries Lift @ 50  & \bfseries Lift @ 100  & \bfseries Lift @ 300  & \bfseries Lift @ 500 \\
        \midrule
    \begin{filecontents*}{lift/complete_resnet_lift.csv}
dataset,model,quality_method,lift_at_10,lift_at_50,lift_at_100,lift_at_300,lift_at_500,lift_at_1000
complete,ResNet-18,Agreement,25.97,33.77,44.16,26.41,17.92,9.48
complete,ResNet-18,Dawid-Skene,27.4,41.1,39.73,25.57,17.81,9.32
complete,ResNet-18,GLAD,89.74,66.67,47.44,18.38,11.03,5.51
complete,ResNet-18,Dawid-Skene with Model,28.17,36.62,38.03,26.29,17.75,9.44
complete,ResNet-18,GLAD with Model,53.33,61.33,46.67,17.78,10.67,5.33
complete,ResNet-18,Empirical Bayes,77.92,49.35,44.16,26.84,18.18,9.48
complete,ResNet-18,Active Label Cleaning,0.0,10.39,12.99,8.66,8.83,5.45
complete,ResNet-18,Label Quality Score,38.96,20.78,19.48,13.42,9.09,6.36
complete,ResNet-18,CROWDLAB,38.96,49.35,44.16,27.71,18.44,9.61
\end{filecontents*}
\csvreader[head to column names]{lift/complete_resnet_lift.csv}{}
    {  \model & \csvcoliii & \csvcoliv & \csvcolv & \csvcolvi & \csvcolvii & \csvcolviii \\} \\[-1em]
        \midrule 
        \begin{filecontents*}{lift/complete_swin_lift.csv}
dataset,model,quality_method,lift_at_10,lift_at_50,lift_at_100,lift_at_300,lift_at_500,lift_at_1000
complete,Swin,Agreement,26.32,34.21,43.42,26.32,17.89,9.47
complete,Swin,Dawid-Skene,41.1,41.1,39.73,25.57,17.81,9.32
complete,Swin,GLAD,89.74,66.67,47.44,18.38,11.03,5.51
complete,Swin,Dawid-Skene with Model,14.93,38.81,37.31,25.87,17.61,9.25
complete,Swin,GLAD with Model,28.17,53.52,46.48,17.37,10.42,5.21
complete,Swin,Empirical Bayes,78.95,50.0,42.11,26.32,17.89,9.34
complete,Swin,Active Label Cleaning,0.0,0.0,5.26,7.46,6.84,5.0
complete,Swin,Label Quality Score,26.32,21.05,21.05,17.11,13.68,8.42
complete,Swin,CROWDLAB,65.79,65.79,48.68,28.07,18.68,9.61
\end{filecontents*}
\csvreader[head to column names]{lift/complete_swin_lift.csv}{} 
           { \model & \csvcoliii & \csvcoliv & \csvcolv & \csvcolvi & \csvcolvii & \csvcolviii \\}
 \\[-1em]
    \bottomrule \\[-0.5em]
\end{tabular}
 \caption{Evaluating the precision of various consensus quality scoring methods on the \emph{Complete} dataset. Lift@$T$ is directly proportional to Precision@$T$, and reports what fraction of the top-$T$ ranked consensus labels are actually incorrect normalized by the fraction of incorrect consensus labels expected for a random set of examples.}
 \label{tab:lift_complete}
\end{table}

\newpage
\section{Model Trained on True CIFAR-10 Labels}
\label{sec:truelabels}

In this section, we investigate how the methods perform when utilizing a highly accurate model. We obtain such a model by training a Swin Transformer on the ground truth labels rather than consenus labels estimated from the given annotations (this would not be possible in real applications). All benchmark results presented in this section are with respect to this unrealistically good classifier.

\begin{figure}[H]
\begin{subfigure}{.5\textwidth}
  \centering
  \includegraphics[width=\linewidth]{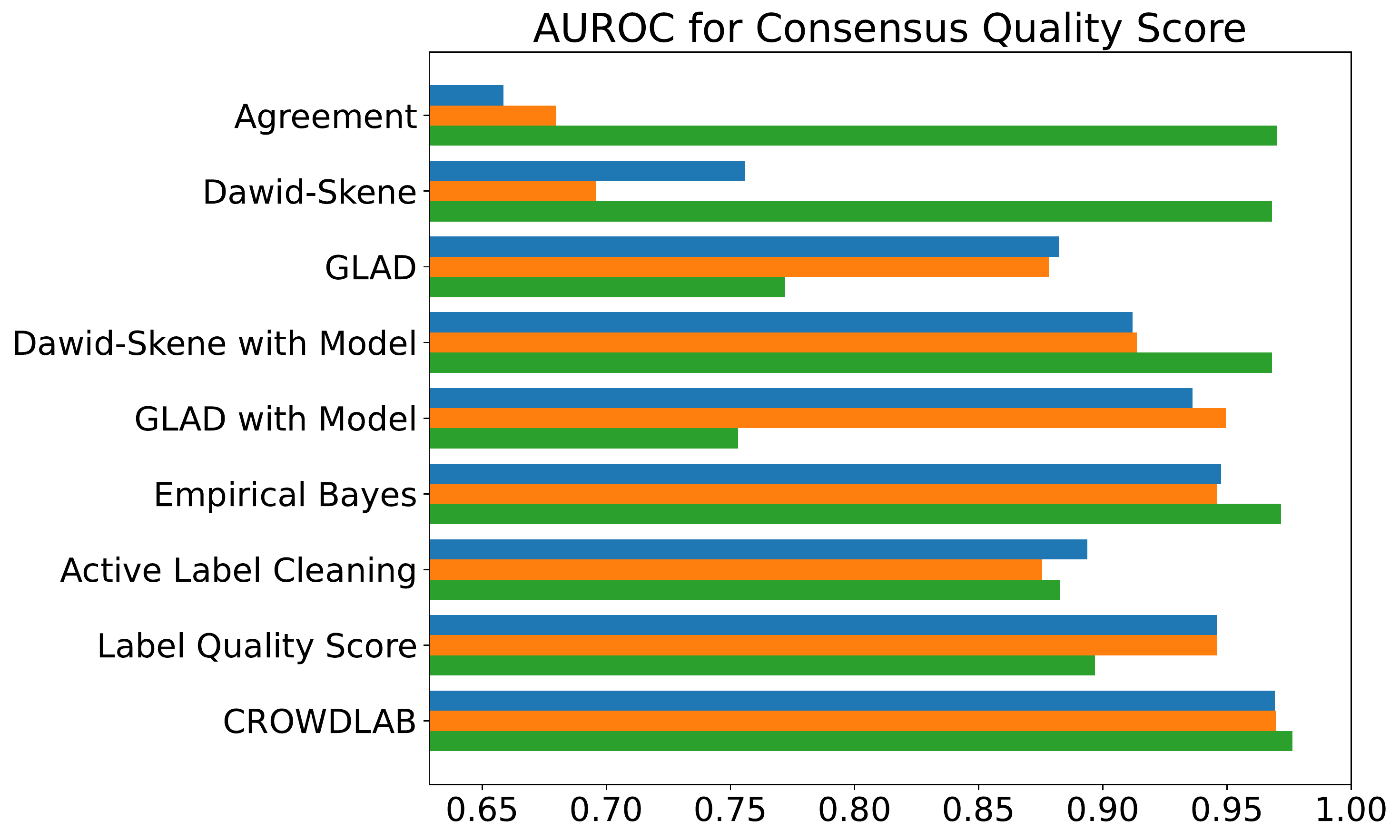} 
  \label{fig:auroc_truelabels}
\end{subfigure}
\begin{subfigure}{.5\textwidth}
  \centering
  \includegraphics[width=\linewidth]{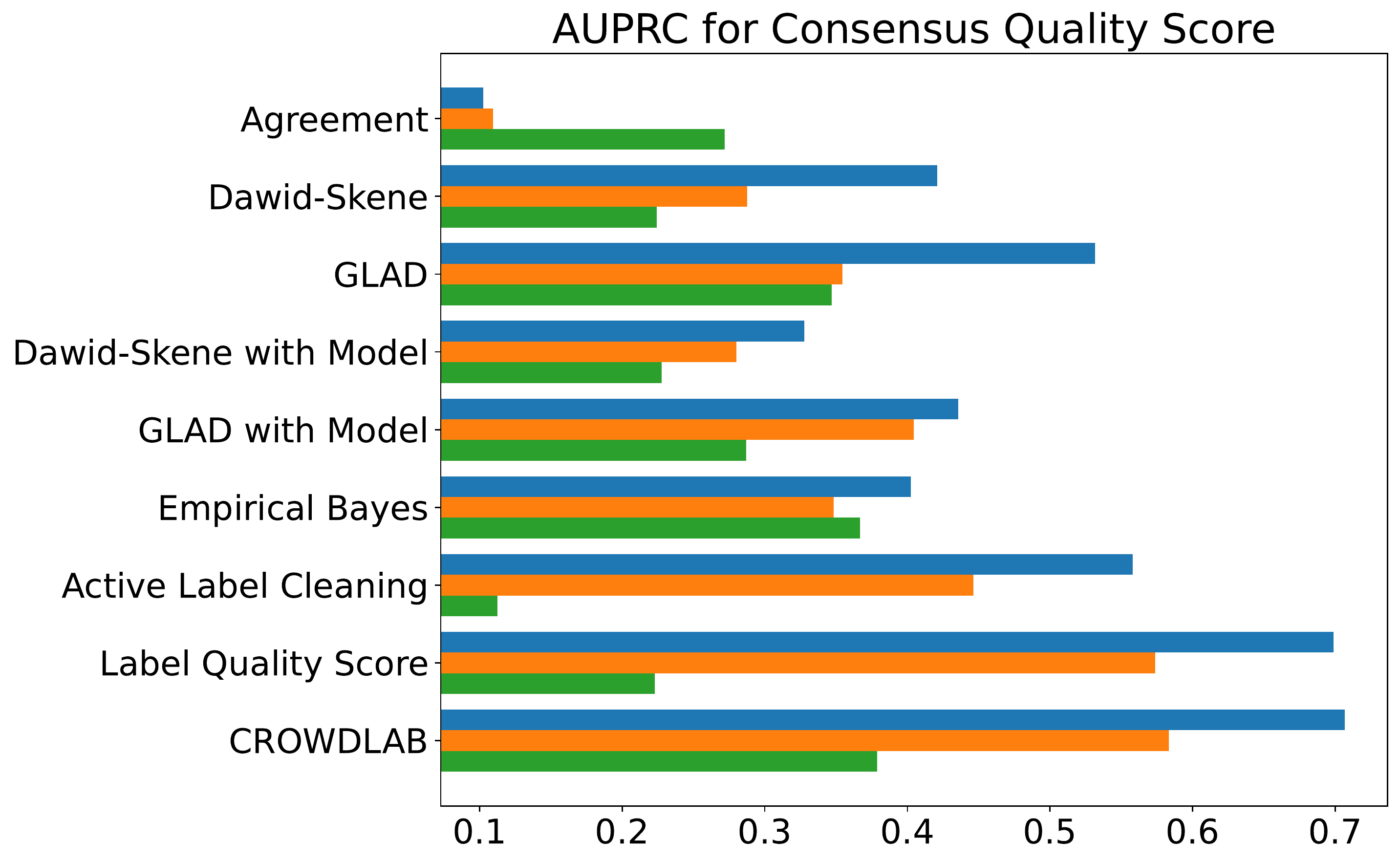}
  \label{fig:auprc_truelabels}
\end{subfigure}
\begin{subfigure}{.49\textwidth}
  \centering
  \includegraphics[width=\linewidth]{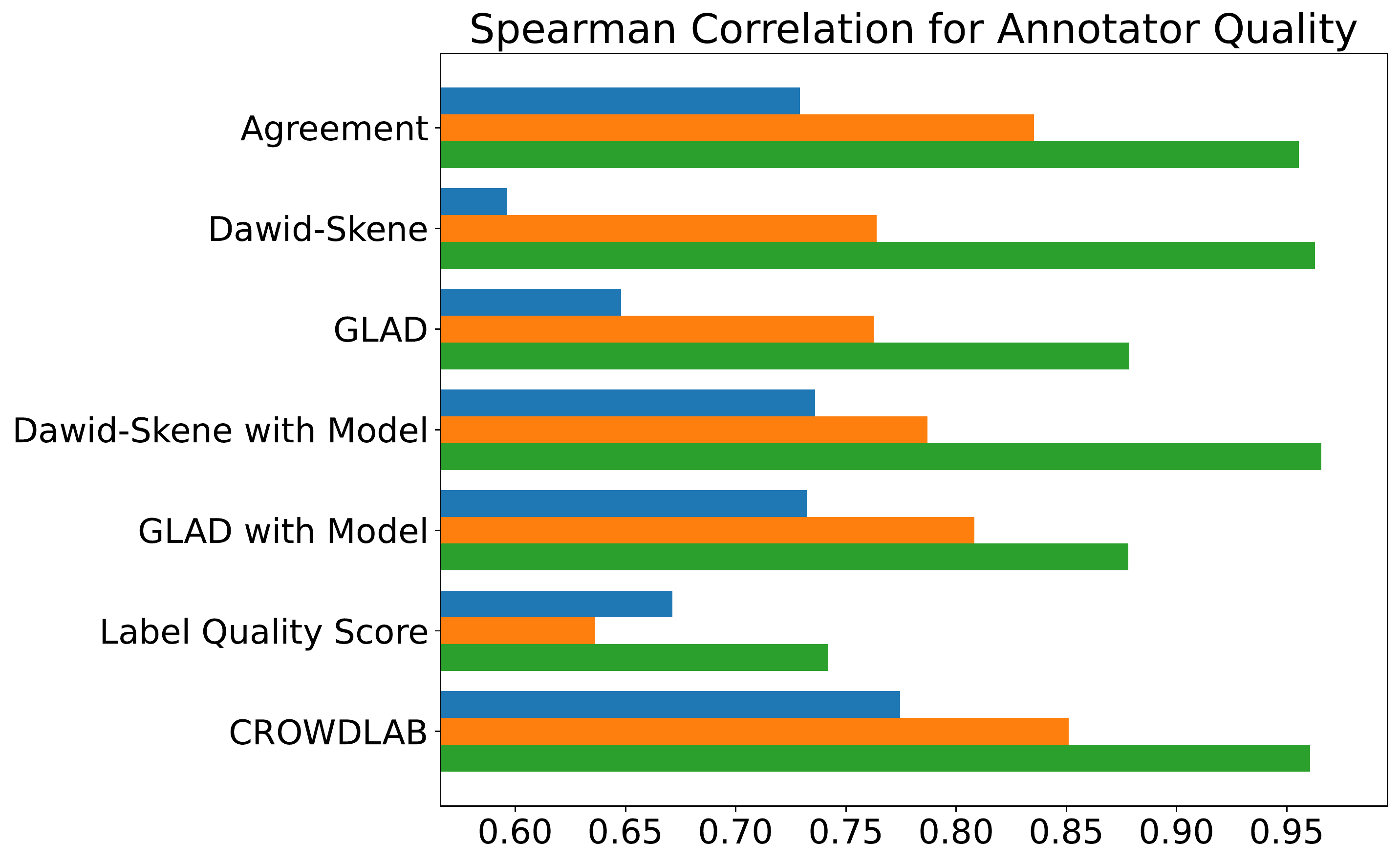}
  \label{fig:spearman_truelabels}
\end{subfigure} 
\begin{subfigure}{.51\textwidth}
  \centering
  \includegraphics[width=\linewidth]{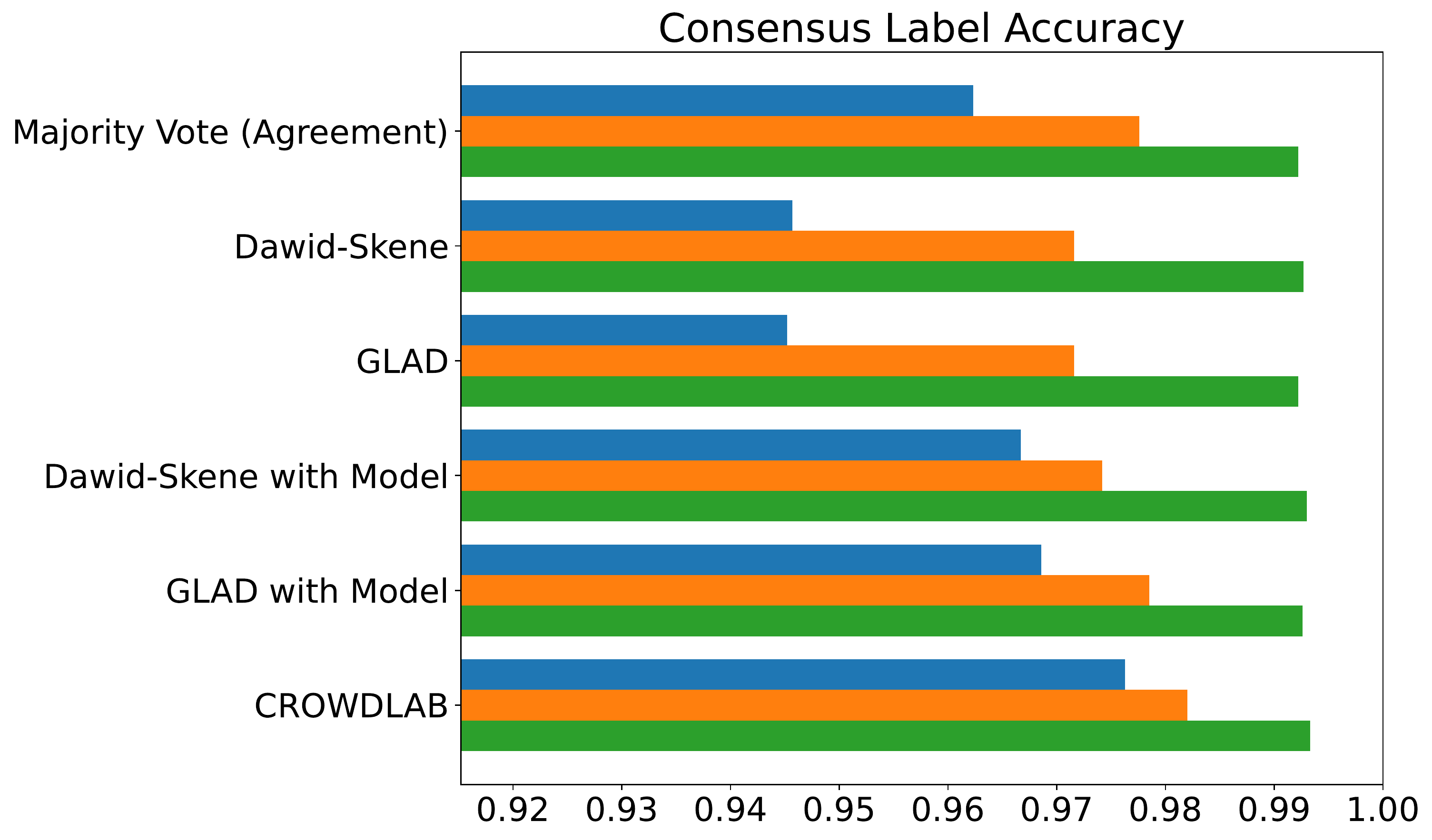}
  \label{fig:accuracy_truelabels}
\end{subfigure}
\begin{subfigure}{\textwidth}
  \centering
  \includegraphics[width=0.4\linewidth]{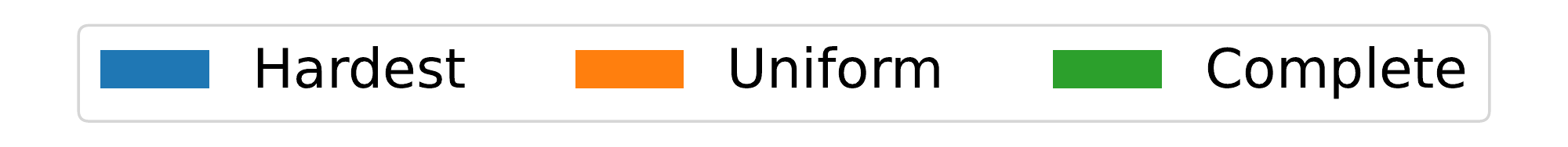} 
  \label{fig:legend_truelabels}
\end{subfigure}
\caption{Benchmarking multi-annotator methods that utilize an  unrealistically good classifier fit to true labels for each dataset.}
\label{fig:truelabels}
\end{figure}

\begin{table}[H]
\centering 
\footnotesize
\begin{tabular}{l c c c c c}
    \toprule 
     \bfseries Quality Method & \bfseries Lift @ 10 & \bfseries Lift @ 50  & \bfseries Lift @ 100  & \bfseries Lift @ 300  & \bfseries Lift @ 500  \\
            \midrule
    \begin{filecontents*}{lift/worst_truelabels_lift.csv}
dataset,model,quality_method,lift_at_10,lift_at_50,lift_at_100,lift_at_300,lift_at_500,lift_at_1000
Hardest,truelabels,Agreement,2.65,3.18,3.98,4.16,3.77,3.74
Hardest,truelabels,Dawid-Skene,14.73,15.47,15.1,11.48,8.95,5.34
Hardest,truelabels,GLAD,14.6,15.69,16.24,14.6,10.07,5.93
Hardest,truelabels,Dawid-Skene with Model,24.02,18.62,15.02,10.11,9.13,6.55
Hardest,truelabels,GLAD with Model,19.11,21.66,20.06,16.45,11.97,8.22
Hardest,truelabels,Empirical Bayes,5.31,7.43,10.34,12.73,12.84,8.33
Hardest,truelabels,Active Label Cleaning,23.87,25.46,25.2,16.18,11.03,6.63
Hardest,truelabels,Label Quality Score,23.87,24.93,24.93,20.07,14.64,8.2
Hardest,truelabels,CROWDLAB,26.53,25.99,24.14,19.19,14.8,8.86
\end{filecontents*}
\csvreader[head to column names]{lift/worst_truelabels_lift.csv}{}
    {  \csvcoliii & \csvcoliv & \csvcolv & \csvcolvi & \csvcolvii & \csvcolviii \\} \\[-1em]
        \bottomrule \\
\end{tabular}
 \caption{Evaluating the lift (i.e.\ precision) of various consensus quality scoring methods on the \emph{Hardest} dataset, here employing our unrealistically good classifier trained with true labels.}
 \label{tab:lift_unrealistic}
\end{table}

\begin{table}[H]
\centering 
\footnotesize
\begin{tabular}{l c c c c c}
    \toprule 
     \bfseries Quality Method & \bfseries Lift @ 10 & \bfseries Lift @ 50  & \bfseries Lift @ 100  & \bfseries Lift @ 300  & \bfseries Lift @ 500  \\
            \midrule
    \begin{filecontents*}{lift/uniform_truelabels_lift.csv}
dataset,model,quality_method,lift_at_10,lift_at_50,lift_at_100,lift_at_300,lift_at_500,lift_at_1000
Uniform,truelabels,Agreement,8.93,10.71,8.48,7.44,6.34,4.33
Uniform,truelabels,Dawid-Skene,28.17,25.35,20.42,12.21,8.73,5.46
Uniform,truelabels,GLAD,31.69,24.65,27.82,13.03,9.23,5.32
Uniform,truelabels,Dawid-Skene with Model,31.01,19.38,14.73,9.95,9.69,7.29
Uniform,truelabels,GLAD with Model,13.95,25.12,23.72,20.78,15.44,8.56
Uniform,truelabels,Empirical Bayes,13.39,19.64,20.98,19.05,14.38,8.12
Uniform,truelabels,Active Label Cleaning,40.18,39.29,32.14,17.11,11.34,6.92
Uniform,truelabels,Label Quality Score,40.18,40.18,36.16,19.79,14.38,8.21
Uniform,truelabels,CROWDLAB,44.64,33.04,34.38,22.32,15.98,9.11
\end{filecontents*}
\csvreader[head to column names]{lift/uniform_truelabels_lift.csv}{}
    {  \csvcoliii & \csvcoliv & \csvcolv & \csvcolvi & \csvcolvii & \csvcolviii \\} \\[-1em]
        \bottomrule \\
\end{tabular}
 \caption{Evaluating the lift (i.e.\ precision) of various consensus quality scoring methods on the \emph{Uniform} dataset, here employing our unrealistically good classifier trained with true labels.}
\end{table}

\begin{table}[H]
\centering 
\footnotesize
\begin{tabular}{l c c c c c}
    \toprule 
     \bfseries Quality Method & \bfseries Lift @ 10 & \bfseries Lift @ 50  & \bfseries Lift @ 100  & \bfseries Lift @ 300  & \bfseries Lift @ 500  \\
            \midrule
    \begin{filecontents*}{lift/complete_truelabels_lift.csv}
dataset,model,quality_method,lift_at_10,lift_at_50,lift_at_100,lift_at_300,lift_at_500,lift_at_1000
Complete,truelabels,Agreement,25.64,35.9,43.59,26.5,17.95,9.49
Complete,truelabels,Dawid-Skene,27.4,38.36,39.73,25.57,17.81,9.32
Complete,truelabels,GLAD,89.74,66.67,47.44,18.38,11.28,5.77
Complete,truelabels,Dawid-Skene with Model,42.86,40.0,38.57,26.67,17.71,9.43
Complete,truelabels,GLAD with Model,40.54,54.05,45.95,17.12,10.54,5.41
Complete,truelabels,Empirical Bayes,89.74,48.72,44.87,26.92,18.21,9.49
Complete,truelabels,Active Label Cleaning,38.46,23.08,19.23,14.53,11.54,7.05
Complete,truelabels,Label Quality Score,64.1,51.28,38.46,18.38,12.56,7.05
Complete,truelabels,CROWDLAB,89.74,61.54,42.31,26.92,18.46,9.62
\end{filecontents*}
\csvreader[head to column names]{lift/complete_truelabels_lift.csv}{}
    {  \csvcoliii & \csvcoliv & \csvcolv & \csvcolvi & \csvcolvii & \csvcolviii \\} \\[-1em]
        \bottomrule \\
\end{tabular}
 \caption{Evaluating the lift (i.e.\ precision) of various consensus quality scoring methods on the \emph{Complete} dataset, here employing our unrealistically good classifier trained with true labels.}
\end{table}

\newpage
\section{Variant of our Method Without Per Annotator Weights}
\label{sec:variants}

Here we present results for a simpler variant of CROWDLAB that we also explored, henceforth called \emph{No Perannotator Weights}.
The two approaches are overall the same, except while CROWDLAB considers each annotator individually and  assigns them a separate weight $w_j$, \emph{No Perannotator Weights} aggregates all the annotators and treats them as one ``average annotator'' to be weighed against the classifier model. 
Details of the \emph{No Perannotator Weights} approach are presented below.

\subsection{Consensus Quality Method}
\label{sec:npw_cqs}

Just as in CROWDLAB, we estimate the quality of consensus labels via the label quality score based on estimated class probabilities. In the \emph{No Perannotator Weights} variant, these probabilities are computed via a slightly different weighted average:
\begin{equation}
    \posterior{i}{NPW} =  \frac{w_{\mathcal{M}} \cdot \probclassifiersim{i} +  w_{\mathcal{A}} \cdot \posteriornoxnotxt{i}{\mathcal{A}} }{ w_{\mathcal{M}} + w_{\mathcal{A}} } 
    \label{eq:cqs_noper}
\end{equation}
where $\displaystyle w_{\mathcal{M}} = w \cdot \frac{1}{n} \sum_i \sqrt{|\mathcal{J}_i|}, \  w_{\mathcal{A}} = (1 - w) \cdot \sqrt{|\mathcal{J}_i|}$ are one weight for the model and one weight applied to all annotators. Both depend on $w$, whose definition follows a similar strategy as used in CROWDLAB for individual annotators, but here applied to their aggregate output.

First let's recall these quantities from Sec. \ref{sec:cqs}: 
$s_j$ represents annotator $j$'s agreement with other annotators who labeled the same examples and is defined in (\ref{eq:sj}), $A_j$ represents the accuracy of each annotator's labels with respect to the majority-vote consensus label for examples with more than one annotation and is defined in (\ref{eq:aj}). 
In this variant, we compute an average annotator accuracy $\bar{A}$ by taking the average of each annotator's accuracy weighted simply by the number of examples each annotator labeled (rather than their estimated trustworthiness).
\begin{equation*}
    \widebar{A} = \frac{\sum_j A_j \cdot |\mathcal{I}_j|}{\sum_j |\mathcal{I}_j|}
\end{equation*}
Let $A_{\mathcal{M}}$ represent the accuracy of the model with respect to the majority-vote consensus labels among examples with more than one annotation, as defined in (\ref{eq:am}). We then choose our weight
$ \displaystyle
w = A_{\mathcal{M}} / ( A_{\mathcal{M}} + \widebar{A})
$ to balance model accuracy vs.\ that of the average annotator.

While CROWDLAB uses a separate class likelihood vector for each annotator, this variant only considers their aggregate class likelihood
\begin{equation*}
    \posteriornoxnotxtk{i}{\mathcal{A}} = \frac{1}{|\mathcal{J}_i|} \sum_{j \in \mathcal{J}_i} P_j  \ \ \ \text{ where } P_j = \begin{cases}
    s_j & \mbox{when } Y_{ij} = k\\
    \frac{1 - s_j}{K - 1} & \mbox{when } Y_{ij} \neq k
\end{cases}
\end{equation*}

\subsection{Annotator Quality Method}

In the \emph{No Perannotator Weights} variant, we score the quality of each annotator via:
\begin{equation*}
    a_j = w \cdot Q_j + (1-w) \cdot A_j
\end{equation*}

Here $Q_j$ the average label quality score of labels given by each annotator, computed via (\ref{eq:qj}) as in CROWDLAB, but here based on class probabilities $\widehat{p}_\text{NPW}$ estimated using \emph{No Perannotator Weights} defined in (\ref{eq:cqs_noper}) in place of $\widehat{p}_\text{CR}$.  $A_j$ and $w$  are defined as above in Sec. \ref{sec:npw_cqs}.

\subsection{Benchmarking CROWDLAB  with/without per annotator weights}

Ignoring the strengths and weakness of each individual annotator when aggregating them is overall detrimental to  CROWDLAB. However the performance reduction due to this modification is surprisingly small, given how important  accounting for annotators' relative quality is stated to be in the crowdsourcing literature \cite{hovy2013learning, karger2011iterative, kara2015modeling, dawidskene, glad}. Rather the key aspects behind the success of CROWDLAB are its careful consideration of: how much to trust the classifier model vs.\ the aggregate annotations along with how many annotations were provided for each example. Studying additional variants of CROWDLAB with either of these two pieces removed produced very poor results in our benchmarks.

\begin{figure}[H]
\begin{subfigure}{.5\textwidth}
  \centering
  \includegraphics[width=\linewidth]{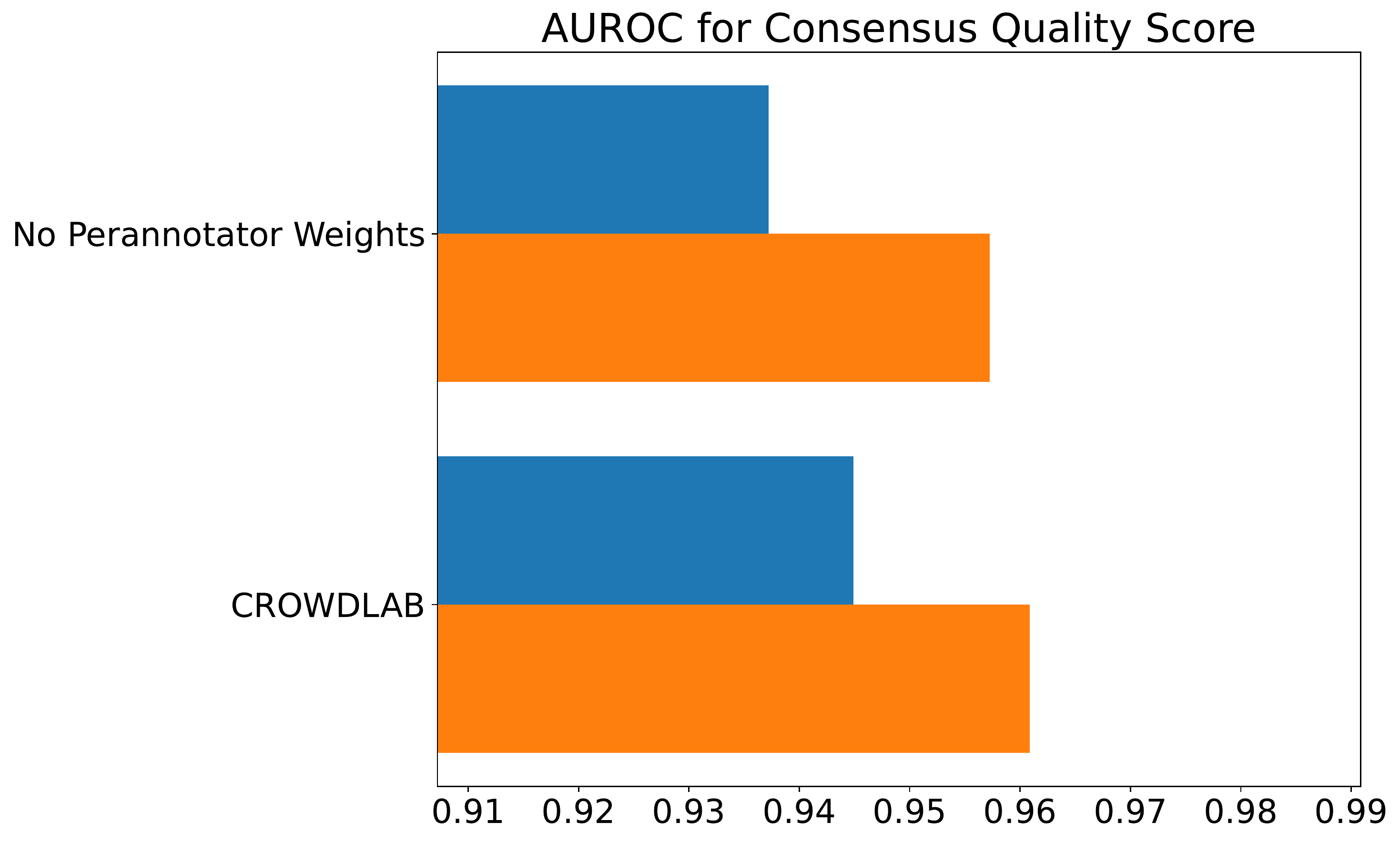} 
  \label{fig:auroc_worst_npw}
\end{subfigure}
\begin{subfigure}{.5\textwidth}
  \centering
  \includegraphics[width=\linewidth]{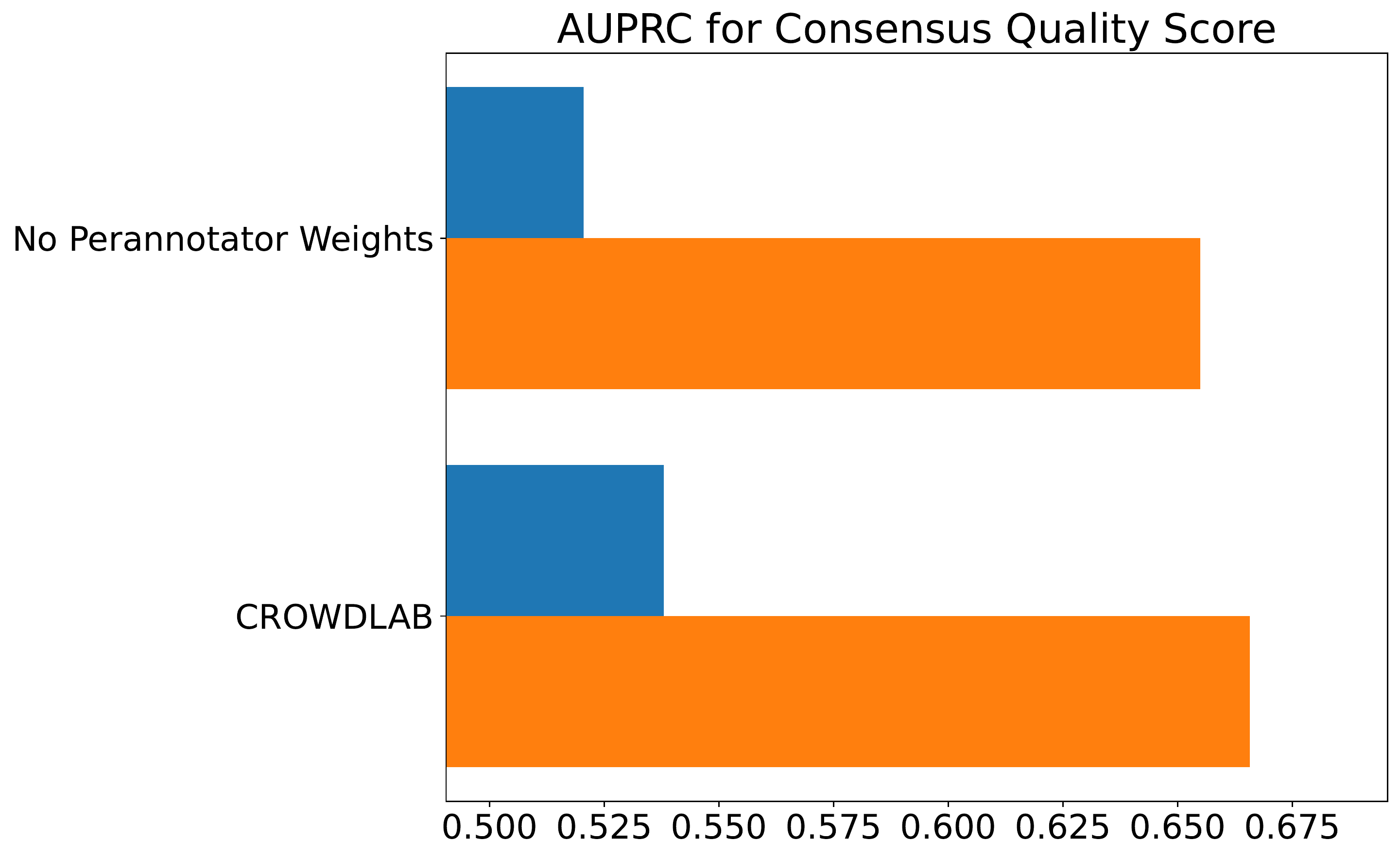} 
  \label{fig:auprc_worst_npw}
\end{subfigure}
\begin{subfigure}{.5\textwidth}
  \centering
  \includegraphics[width=\linewidth]{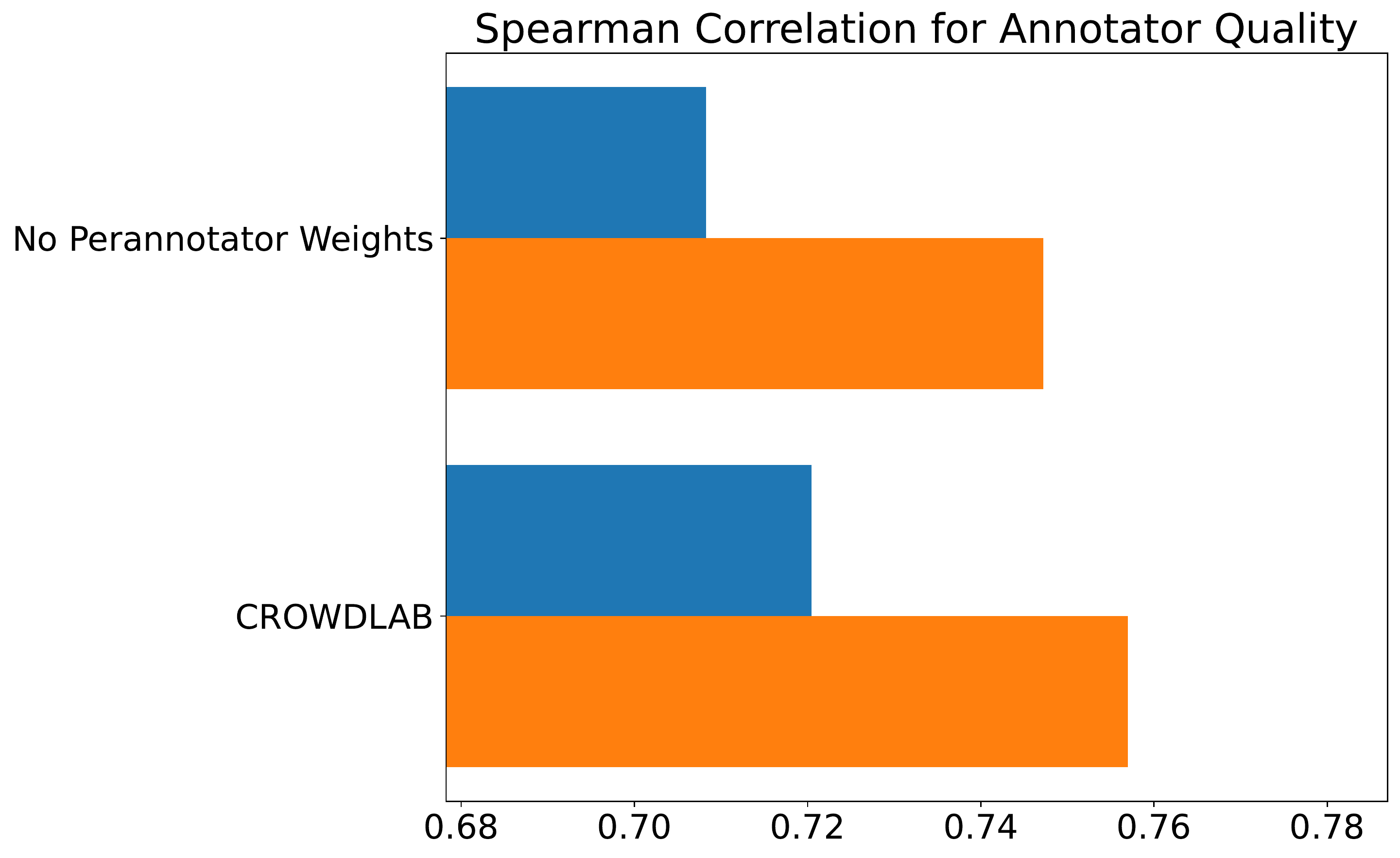}
  \label{fig:spearman_worst_npw}
\end{subfigure} 
\begin{subfigure}{.5\textwidth}
  \centering
  \hspace*{7mm} \includegraphics[width=\linewidth]{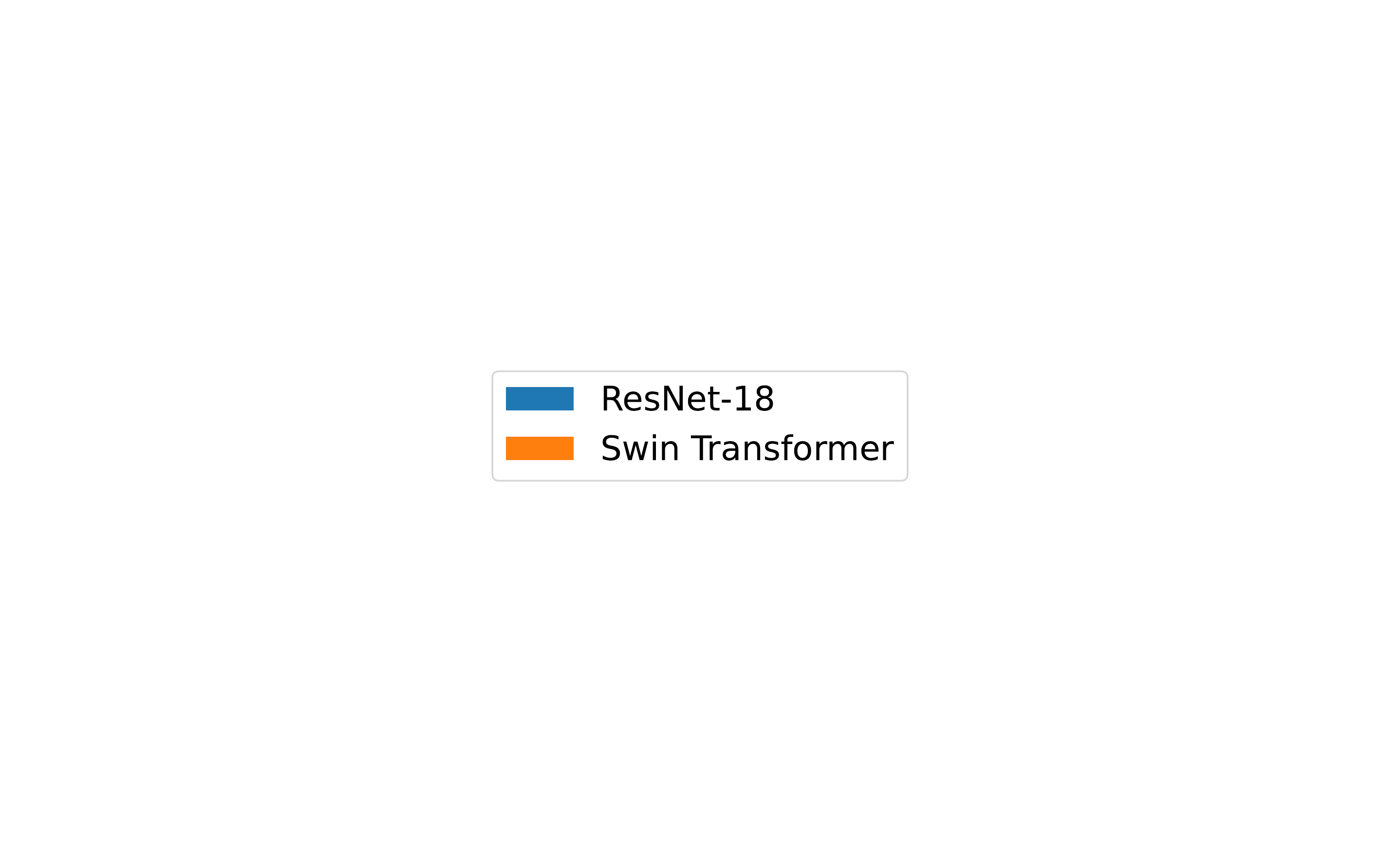} 
  \label{fig:legend_worst_npw}
\end{subfigure}
\caption{Benchmarking CROWDLAB with/without per annotator weights on the \emph{Hardest} dataset.}
\label{fig:noperannotator}
\end{figure}

\vspace*{5mm}
\begin{table}[H]
\centering 
\footnotesize
\begin{tabular}{l l c c c c c}
    \toprule 
    \bfseries Model & \bfseries Quality Method & \bfseries Lift @ 10 & \bfseries Lift @ 50  & \bfseries Lift @ 100  & \bfseries Lift @ 300  & \bfseries Lift @ 500  \\
                \midrule
    \begin{filecontents*}{lift/worst_resnet_lift_npw.csv}
dataset,model,quality_method,lift_at_10,lift_at_50,lift_at_100,lift_at_300,lift_at_500,lift_at_1000
worst_annotators,ResNet-18,No Perannotator Weights,24.33,21.9,18.25,13.95,11.92,7.54
worst_annotators,ResNet-18,CROWDLAB,24.33,22.38,17.76,14.27,11.82,7.76
\end{filecontents*}
\csvreader[head to column names]{lift/worst_resnet_lift_npw.csv}{}
    {\model & \csvcoliii & \csvcoliv & \csvcolv & \csvcolvi & \csvcolvii & \csvcolviii \\} \\[-1em]
                    \midrule
     \begin{filecontents*}{lift/worst_swin_lift_npw.csv}
dataset,model,quality_method,lift_at_10,lift_at_50,lift_at_100,lift_at_300,lift_at_500,lift_at_1000
worst_annotators,Swin,No Perannotator Weights,25.13,23.62,20.85,17.84,14.02,8.49
worst_annotators,Swin,CROWDLAB,25.13,24.62,20.85,17.76,13.82,8.82
\end{filecontents*}
\csvreader[head to column names]{lift/worst_swin_lift_npw.csv}{} 
    {\model & \csvcoliii & \csvcoliv & \csvcolv & \csvcolvi & \csvcolvii & \csvcolviii \\} \\[-1em]
        \bottomrule \\
\end{tabular}
 \caption{Evaluating the precision of CROWDLAB consensus quality scores with/without per annotator weights  on the \emph{Hardest} dataset. Lift@$T$ is directly proportional to Precision@$T$, and reports what fraction of the top-$T$ ranked consensus labels are actually incorrect, normalized by the fraction of incorrect consensus labels expected for a random set of examples.}
\end{table}

\end{document}